\documentclass[onecolumn]{article}
\usepackage{bagrow}

\usepackage{fullpage}
\usepackage{amsmath,amssymb,amsfonts}
\usepackage{graphicx}
\usepackage{hyperref}
\usepackage[numbers]{natbib}
\usepackage{booktabs}
\usepackage{rotating}  %
\usepackage{multirow}
\usepackage{bbm}
\usepackage{bm}
\usepackage{enumitem}

\usepackage{geometry}
\usepackage{caption}
\usepackage{enumitem}
\setlist[itemize]{itemsep=0pt}

\usepackage{xcolor}  %

\DeclareMathOperator*{\argmax}{arg\,max}

\usepackage{xcolor}
\usepackage{framed}
\colorlet{shadecolor}{yellow!30}  %

\renewcommand{\vec}[1]{\ensuremath{\mathbf{#1}}}

\newcommand{\lzero}{\ell_0}
\newcommand{\lone}{\ell_1}
\newcommand{\lznorm}[1]{\|#1\|_0}
\title{Optimized Architectures for Kolmogorov--Arnold Networks}

\author[1,2,*]{James Bagrow}
\author[3,2]{Josh Bongard}
\affil[1]{Mathematics \& Statistics, University of Vermont, Burlington, VT, United States }
\affil[2]{Vermont Complex Systems Center, University of Vermont, Burlington, VT, United States}
\affil[3]{Computer Science, University of Vermont, Burlington, VT, United States}
\affil[*]{\corrauthinfo{james.bagrow@uvm.edu}{bagrow.com}
}

\date{\today}

\begin{document}

\maketitle
\begin{abstract}\small
Efforts to improve Kolmogorov--Arnold networks (KANs) with architectural enhancements have been stymied by the complexity those enhancements bring, undermining the interpretability that makes KANs attractive in the first place.
Here we study overprovisioned architectures combined with sparsification, deep supervision, and depth selection, to learn compact, interpretable KANs without sacrificing accuracy.
Crucially, we focus on differentiable mechanisms under a principled minimum description length objective, jointly optimizing activations, structure, and depth end-to-end.
Experiments across function approximation benchmarks, dynamical systems forecasting, and real-world prediction tasks demonstrate that sparsification alone is insufficient, but the combination with depth selection achieves competitive or superior accuracy while discovering substantially smaller models.
The result is a principled path toward models that are both more expressive and more interpretable, addressing a key tension in scientific machine learning.
\end{abstract}


\vspace{1em}

\section{Introduction}

Deep learning has transformed scientific modeling~\cite{jumper2021highly,karniadakis2021physics,wang2023scientific}, but its advances often work against interpretability~\cite{rudin2019stop,guidotti2018survey}.
Skip connections~\cite{he2016deep}, DenseNet blocks~\cite{huang2017densely}, and deeper architectures improve accuracy by adding complexity. 
This complexity makes models harder to understand.
For scientific applications, where insight matters as much as prediction, the tension is acute.

Sparsification offers a way out~\cite{hastie2015statistical,lecun1989optimal,sindy2016,louizos2018learning}.
Starting from an overprovisioned model, learn which components can be removed without sacrificing (too much) accuracy. 
When sparsification is differentiable, structure and parameters can be learned jointly~\cite{louizos2018learning,liu2018darts}, avoiding the expense of traditional, discrete architecture search~\cite{zoph2016neural,elsken2019neural}.

\begin{figure*}[t]
     \centering
     \includegraphics[width=0.8\textwidth]{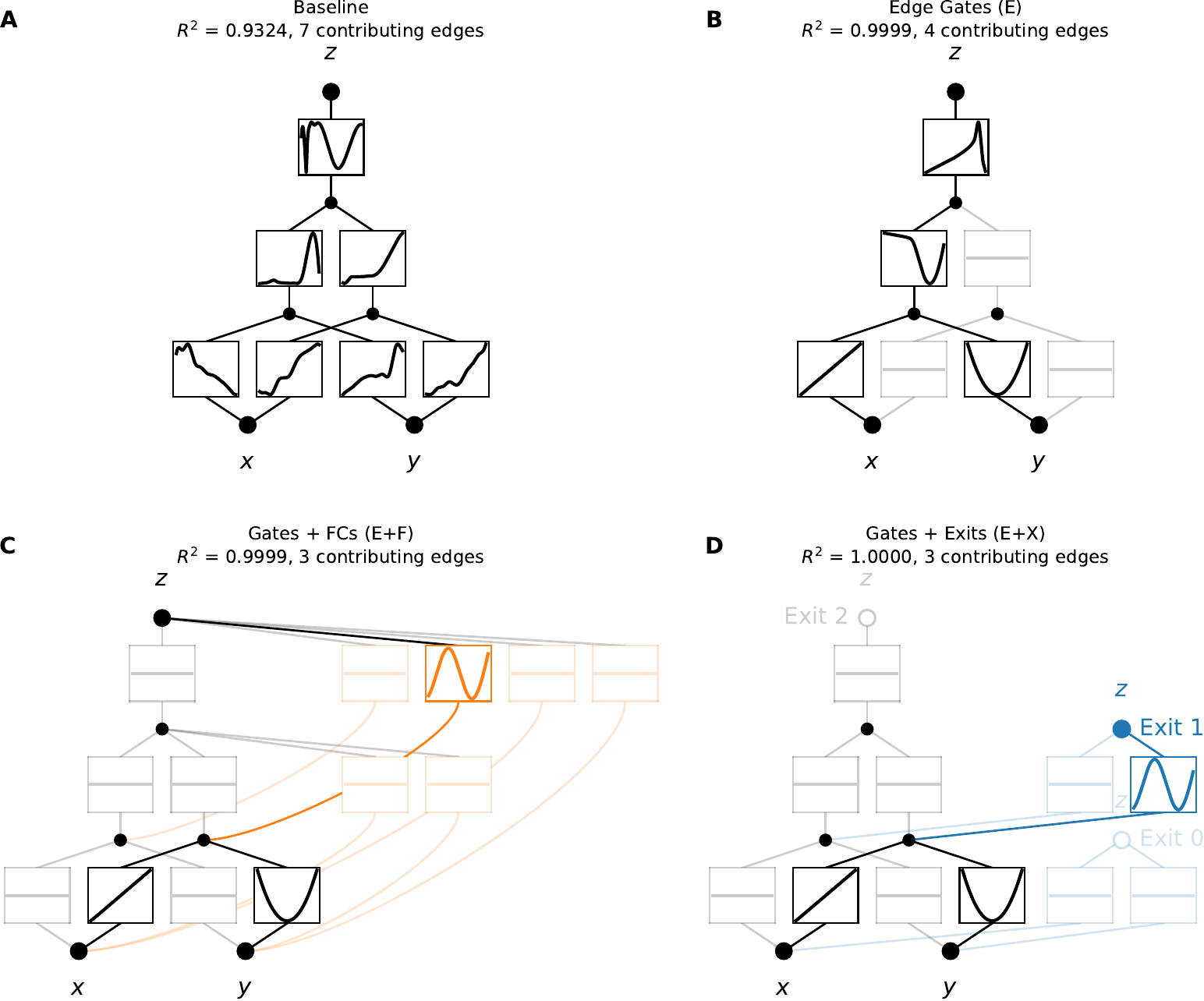}
     \caption{Learning $z = \sin\left(x + y^2\right)$ with overprovisioned KANs (architecture $[2,2,1,1]$). 
     Forward connections are shown in orange and exit gates in blue.}
     \label{fig:anecdote}
 \end{figure*}

Recently, Kolmogorov--Arnold networks (KANs) have shown a unique combination of predictive accuracy and interpretability, making them useful for modeling scientific systems~\cite{liu2025kan,liu2024kan2.0}.
Standard networks learn weights while KANs learn univariate activation functions (Fig.~\ref{fig:anecdote}A), a shift that makes individual components inspectable.
But KANs face the same tradeoff as standard networks. 
Overprovisioning improves expressiveness at the expense of interpretability. 
Larger models become hard to interpret, and thus parsimony, as well as accuracy, is important.
Again, sparsification is the way out.

In this paper, we show that KANs can be made more expressive (Fig.~\ref{fig:anecdote}B-D) without sacrificing, and often improving, interpretability. 
We combine three mechanisms: 
differentiable \textit{edge gates} for sparsification,
DenseNet-style \textit{forward connections},
and learnable layer-wise \textit{exit gates}.
A minimum description length (MDL) objective provides principled guidance for the tradeoff between accuracy and complexity, allowing joint learning of activations, sparsity, and depth.
We evaluate these mechanisms using a $2 \times 2 \times 2$ design across function approximation benchmarks, dynamical systems modeling, and real-world prediction tasks.
Our key finding is that edge-level sparsification is necessary but not sufficient; however, when combined with mechanisms for depth selection and deep supervision, overprovisioned KANs can be pruned during training to compact, accurate models.

The rest of this paper is organized as follows.
Section~\ref{sec:background} provides background on KANs, DenseNet-style forward connections, multi-exit networks, differentiable $\ell_0$ regularization, and the Gumbel-Softmax relaxation used for exit selection.
Section~\ref{sec:okan} presents our gated multi-exit architecture, an MDL-motivated objective function, and the training procedure.
Section~\ref{sec:results} evaluates all eight conditions of the $2\times 2\times 2$ factorial across symbolic benchmarks, dynamical systems, and real-world datasets.
We conclude with a discussion in Sec.~\ref{sec:discussion}.

\section{Background}
\label{sec:background}

\subsection{Kolmogorov--Arnold Networks}
\label{subsec:background:kans}

Kolmogorov--Arnold Networks (KANs), motivated by the Kolmogorov--Arnold Representation Theorem~\cite{kolmogorov1961representation,arnold2009functions,kolmogorov1957representations}, 
consist of $L$ layers with shapes $[n_0, n_1, \ldots, n_L]$. 
The layer update is given by:
\begin{equation}
    x^{(\ell+1)}_j = \sum_{i=1}^{n_\ell} \phi_{\ell ij}\left(x^{(\ell)}_i\right)
    \label{eqn:kan-layer-update-j}
\end{equation}
where $\phi_{\ell ij}: \mathbb{R} \to \mathbb{R}$ is a learnable univariate activation function associated with the edge from neuron $i$ in layer $\ell$ to neuron $j$ in layer $\ell+1$.
In matrix form, the layer update becomes $\vec{x}^{(\ell+1)} = \Phi_\ell(\vec{x}^{(\ell)})$, where $\Phi_\ell$ is the functional matrix containing the $\phi_{\ell ij}$'s connecting layers $\ell$ and $\ell+1$, and $\vec{x}^{(0)} := \vec{x}$ is the network input.
Composing the layers yields the full network:
\begin{equation}
    \text{KAN}(\vec{x}) = \left(\Phi_{L-1} \circ \Phi_{L-2} \circ \cdots \circ \Phi_1 \circ \Phi_0\right)(\vec{x}).
    \label{eqn:kan-full-composition}
\end{equation}
Thus the KAN architecture consists of fully-connected layers of activation functions joined by summation nodes, and the architecture is defined by the shape vector $[n_0=q, n_1, \ldots, n_L=p]$, where $q$ and $p$ are the dimensions of the input and output, respectively. 
While summation nodes are theoretically sufficient to represent any function including multiplication, KAN 2.0~\cite{liu2024kan2.0} introduces explicit multiplication nodes alongside summations that compute products of incoming activations, yielding more compact networks for multiplicative functions.

Many parameterizations of $\phi_{\ell ij}$ have been proposed (see, e.g., \cite{li2024fastkan,fourierKAN,reinhardt2024sinekan,bozorgasl2405wav,sidharth2024chebyshev}); here, to make our results as comparable to prior efforts, we use the original B-spline formulation~\cite{liu2025kan,liu2024kan2.0}, which combines a B-spline with a fixed, nonpolynomial base function $b(x)$, typically SiLU ($b(x) = x \sigma(x)$):
\begin{equation}
    \phi(x) = w_b b(x) + w_s \sum_{m=1}^{G+K} c_{m} B_m(x),
\end{equation}
where $w_b$ and $w_s$ are learnable scale parameters, $c_{m}$ are learnable spline coefficients, and $B_m$ are B-spline basis functions of order $K$ over $G$ grid intervals. 
The base function $b(x)$ ensures that \eqref{eqn:kan-full-composition} does not collapse to a high-order polynomial, which would otherwise result from the polynomial B-splines alone.

When KANs were introduced, regularizations of various forms were proposed, mostly to prevent overfitting of the spline terms~\cite{liu2025kan,liu2024kan2.0}. 
\citeauthor{liu2025kan} also introduced pruning of the network as a sparsification technique~\cite{liu2025kan,liu2024kan2.0}. 
However, this approach pruned activation functions after training, in a post-hoc optimization step. 
In contrast, in this work we apply a differentiable sparsification process that learns both the KAN parameters and the sparsity (via gating terms) jointly, during training.

\subsection{DenseNet Forward Connections}
\label{subsec:background:densenet-forward-connections}

\citeauthor{huang2017densely}~\cite{huang2017densely} introduce dense blocks, where each layer in the block is connected to every subsequent layer in that block. 
These \emph{forward connections} (FCs) allow inputs and learned features to be accessible to later layers, operating in a manner similar to ResNet-style skip connections~\cite{he2016deep} but accumulating across layers. 
They also provide deep supervision~\cite{lee2015deeply}, allowing training gradients to flow directly from later layers to earlier ones.
Similar to our work here, forward connections have been used successfully for other forms of neural network-based symbolic regression~\cite{10253952}.
When combined with sparsification, FCs can be used for architecture search, as earlier features and inputs are transported directly to the output layer, bypassing the trunk, effectively regulating the depth of the network.

With FCs, the KAN layer update, Eq.~\eqref{eqn:kan-layer-update-j}, becomes
\begin{equation}
\vec{x}^{(\ell+1)} = \Phi_\ell\left(\left[\vec{x}^{(0)}, \vec{x}^{(1)}, \ldots, \vec{x}^{(\ell)}\right]\right),
\end{equation}
where the input to layer $\ell$ is the concatenation of the network input along with all previous layer outputs.
The cumulative nature of FCs can result in a potentially large number of activation functions. 
With so many activation functions, the KAN becomes more difficult to train but more importantly, more difficult to interpret.
This motivates the need for a sparsification mechanism (Sec.~\ref{subsec:background:diff-l0-sparse}) that can prune unnecessary connections while retaining the benefits of FCs.

\subsection{Multi-exit Networks}
\label{subsec:background:multi-exit-networks}

Multi-exit and early-exit networks are a class of architectures where some or all layers have attached a separate output, or exit head~\cite{scardapane2020should,10.1145/3469116.3470012,10.1145/3698767}, enabling fast inference by exiting early on simpler inputs.
Beyond speed, multi-exits provide another benefit: they allow deep supervision. Because each layer has its own output, backpropagation will inject gradients directly into all layers at once, similar to forward connections but via a different architecture.

In the context of Kolmogorov--Arnold networks,
prior work introduced multi-exit KANs~\cite{Bagrow_2025}.
A multi-exit KAN augments each layer of width $n_\ell$ in a standard KAN with an  exit head of shape $[n_\ell, p]$.
These additional outputs are combined in a weighted loss, with weights either prechosen as hyperparameters or learned during training. 
For a network with $L$ layers (i.e., $L$ exits), and outputs $\hat{y}^{(0)}, \hat{y}^{(1)}, \ldots, \hat{y}^{(L-1)}$, the total loss is
\begin{equation}
\mathcal{L}_\text{multi-exit} = \sum_{k=0}^{L-1} w_k \mathcal{L}_k,
\end{equation}
where $\mathcal{L}_k$ is the individual loss for exit $k$, such as MSE, $ \mathcal{L}_k = \sum_i (\hat{y}_i^{(k)}-y_i)^2/n$, and $w_k$ is the weight for exit $k$. %
In addition to deep supervision, if training is structured to learn which exit to focus on (meaning the $\{w_k\}$ become learnable parameters), multi-exit KANs can also help with architectural search by estimating the appropriate depth of the network.
An inductive bias favoring parsimony can be introduced if learning is designed to favor earlier layers over later ones, which can be helpful with interpretable models such as KANs.

\subsection{Differentiable \texorpdfstring{$\lzero$}{l0} Regularization}
\label{subsec:background:diff-l0-sparse}

A classic approach to sparsifying an overprovisioned model is to use the $\lone$-norm as a proxy for the $\lzero$-norm, as was famously pioneered by LASSO~\cite{10.1111/j.2517-6161.1996.tb02080.x}.
However, LASSO achieves sparse selection through the closed-form proximal update to the soft-thresholding operator, but in the context of a gradient-based optimization algorithm, $\lone$ regularization alone would encourage small parameter values, not values of zero.

\citeauthor{louizos2018learning}~\cite{louizos2018learning} introduce a continuous relaxation of an $\lzero$ regularization term, which encourages sparsity in neural network weights (or any predictive model parameters).  
Their approach possesses two crucial properties: (1) it is differentiable and allows for end-to-end learning, and (2) it encourages true sparsity by setting parameters to exactly zero.

Consider training a predictive model $\vec{y} = f(\vec{x}; \bm{\theta})$, where $\vec{y} \in \mathbb{R}^{n \times p}$ is the target, $\vec{x} \in \mathbb{R}^{n \times q}$ is the input, and $\bm{\theta}$ are the model parameters.
Such a model can be trained by learning the $\bm{\theta}$ that minimizes a loss function regularized by an $\lzero$ penalty that sparsifies $\bm{\theta}$:
\begin{equation}
\mathcal{L}(\bm\theta) = \frac{1}{n}\sum_{i=1}^{n}\mathcal{L}_\text{data}\big(f(\vec{x}_i; \bm{\theta}), \vec{y}_i\big) + \beta \lznorm{\bm{\theta}}, \quad \lznorm{\bm\theta} = \sum_{j}^{|\bm\theta|} \mathbb{I}[\theta_j \neq 0].
\end{equation}
Now, use a binary gate $z_j \in \{0,1\}$ to reparameterize each $\theta_j \gets \theta_j z_j$ ($\bm\theta \gets \bm\theta \odot\vec{z}$) such that $\lznorm{\bm\theta} = \sum_j z_j$, where $\odot$ denotes elementwise multiplication.
We can then learn the $\bm{\theta},\vec{z}$ that minimizes the loss. 
Unfortunately, introducing binary gates discretizes the parameter space and makes the problem non-differentiable.

To make this sparse learning problem differentiable, \citeauthor{louizos2018learning} introduce a continuous relaxation of $\vec{z}$ that is then clamped to $[0,1]$. 
Specifically, for each gate $j$, introduce a learnable parameter $\alpha_j \in \mathbb{R}$ and define a stochastic binary variable:
\begin{equation}
s_j = \sigma\left(\frac{\log u - \log(1-u) + \alpha_j}{\tau}\right), \quad u \sim \text{Uniform}(0,1),
\end{equation}
where $\sigma(\cdot)$ is the sigmoid function and $\tau$ is a temperature hyperparameter. 
Next, stretch and clamp $s_j$:
\begin{equation}
\bar{s}_j = s_j(\zeta - \gamma) + \gamma,
\end{equation}
\begin{equation}
\tilde{z}_j = \min(1, \max(0, \bar{s}_j)).
\end{equation}
To train the model with a gradient-based method such as Adam~\cite{kingma2014adam}, use the reparameterization trick to differentiably sample $\{z_j\}$ and compute gradient updates to $\{\alpha_j\}$.
For the regularization term, the number of open gates,
the probability that $j$ is open has a closed form~\cite{louizos2018learning},
\begin{equation}
\mathbb{E}[\tilde{z}_j] = \sigma\left(\alpha_j - \tau \log \frac{-\gamma}{\zeta}\right),
\label{eqn:expected-z}
\end{equation}
and the complexity loss $\mathcal{L}_C := \sum_j \mathbb{E}[\tilde{z}_j]$ becomes our proxy for $\lznorm{\bm\theta}$ in $\mathcal{L}(\bm\theta)$.
Finally, at inference, use the following estimator for the trained parameters $\bm\theta^{*}$:

\begin{equation}
\hat{z}_j = \min\left(1, \max\left(0, \sigma(\alpha_j)(\zeta - \gamma) + \gamma\right)\right), \qquad \bm\theta^* \gets \bm{\theta}^* \odot \hat{\vec{z}}.
\label{eqn:louizos-test-estimator}
\end{equation}

This form of differentiable sparsity has been successfully applied to neural network architectures across areas~\cite{louizos2018learning,gale2019state}. 
It has also been used to optimize Equation Learner Networks (EQLs)~\cite{martius2016extrapolation,pmlr-v80-sahoo18a}, a fixed-activation function approach to using deep learning for symbolic regression~\cite{kim2020integration,10253952}.
However, it has also been remarked that this approach can exhibit problematic variance, particularly in large networks, and that its hyperparameters can be difficult to tune~\cite{gale2019state,hoefler2021sparsity}.

\subsection{Differentiable Categorical Selection}
\label{subsec:background:gumbel-softmax}

The differentiable gates of Sec.~\ref{subsec:background:diff-l0-sparse} implement independent binary variables, but a different approach is needed for a categorical variable, where choices are mutually exclusive.
To draw samples from a categorical distribution with probabilities $\pi_k$, $k=1,...,K$, we can use the Gumbel-Max trick~\cite{gumbel1954statistical,maddison2014sampling}, $k^{*} = \argmax_k \left(\log \pi_k + g_k\right)$, where $g_k \sim \text{Gumbel}(0,1)$. However, the argmax is not differentiable, so the Gumbel-Softmax relaxation~\cite{jang2017categorical,maddison2017the} replaces it with a softmax function at temperature $\tau$:
\begin{equation}
\eta_k = \text{softmax}\left(\frac{\log \pi_k + g_k}{\tau}\right),
\end{equation}
where $\text{softmax}(x_k) = e^{x_k} / \sum_{j} e^{x_j}$.
The resulting sample $\bm{\eta}$ lies on the $(K{-}1)$-simplex, continuously approximating a one-hot vector, and has a closed-form density called the Concrete distribution~\cite{maddison2017the}, which converges to the categorical distribution with probabilities $\{\pi_k\}$ as $\tau \to 0$.
For training, Jang \emph{et al.}~\cite{jang2017categorical} show that annealing $\tau$ from a high to low value works well across many different annealing schedules. 
Maddison \emph{et al.}~\cite{maddison2017the} show that $\tau \leq (K-1)^{-1}$ guarantees no interior modes on the simplex, ensuring the density concentrates near one-hot vectors. %

\section{Optimizing KAN architectures} %
\label{sec:okan}

To optimize the architecture of a KAN, we introduce an overprovisioned, deeply supervised architecture (Secs.~\ref{subsec:background:densenet-forward-connections} and \ref{subsec:background:multi-exit-networks}) along with a sparsification mechanism (Sec.~\ref{subsec:background:diff-l0-sparse}).  
A principled loss function based on minimum description length allows for optimization.

\subsection{Gated multi-exit architecture}
\label{subsec:gated-architecture}

To enhance the KAN architecture's flexibility, we add forward connections (FCs) between all layers including the inputs and we add exit heads at each layer of the trunk.
Both provide deep supervision where gradients can flow directly to earlier layers and can help the network better represent complex functions and complex compositions of functions.
However, these additions overprovision the network, increasing, potentially greatly, the number of activation functions.

Gating terms address this by allowing the network to sparsify and remove unnecessary activation functions (edges), summation/multiplication units (nodes) and exit heads.
In terms of architecture search, this sparsification also provides depth selection.
Choosing to gate out all but a single exit head removes all the layers in the network trunk after that exit, reducing the learned model's effective depth. 
FCs may also provide depth selection if coupled with sparsifying gates, as they make inputs and early features accessible to the final output. 
If a learned network gates out most of the trunk in favor of the final FCs, this would be equivalent to a network compression that eliminates extraneous layers.

We consider three types of gates, gates on activation functions/edges (\textbf{egates}),  gates on units/nodes (\textbf{ngates}), and a single (categorical) gate across exit heads (\textbf{xgates}). 
Specifically, for each activation function (edge) $\phi_{\ell i j}(x) \gets \phi_{\ell i j}(x) z_{\ell i j}$, where egate $z_{\ell i j} \in \{0,1\}$ is relaxed and learned as described in Sec.~\ref{subsec:background:diff-l0-sparse}.
Likewise, each summation unit (node) can be gated with a corresponding ngate, yielding the new layer update terms:
\begin{equation}
    x^{(\ell+1)}_j = z_{\ell+1,j} \sum_{i=1}^{n_\ell} z_{\ell ij} \phi_{\ell ij}\left(x^{(\ell)}_i\right),
    \label{eqn:gated-layer-update}
\end{equation}
where $z_{\ell+1,j}$ is the ngate on unit $j$ in layer $\ell+1$\footnote{In theory we could include gates on inputs and implement variable selection; we save this for future work.}.
(We distinguish between egates and ngates using the number of indices.)
Multiplication units are gated in the same way.
In matrix form, this becomes $\vec{x}^{(\ell+1)} = \vec{z}_{\ell+1}^{(n)} \odot \left( \Phi_\ell \odot \vec{Z}_\ell^{(e)} \right) (\vec{x}^{(\ell)})$, with $\vec{z}^{(n)}$ collecting the ngates and $\vec{Z}^{(e)}$ the egates.
Node gates provide structured or group sparsity~\cite{yuan2006model,10.1214/09-AOS778}, which may be beneficial for KANs, although they are a special case of egates.
Finally, one differentiably-sampled categorical variable (Sec.~\ref{subsec:background:gumbel-softmax}) learns the exit gate probabilities, $\pi_k$, that the layer $k$ exit is used as the final network output.
(All edges in a gated-out exit head are turned off, but edges inside the active exit head can be gated or not via the egates.)
This provides direct depth selection, which has been shown to benefit KANs~\cite{Bagrow_2025}, unlike the indirect selection available via FCs and egates.

\subsection{Learning objective}
\label{subsec:learning-objective}

To learn KANs that are both accurate and parsimonious, following~\citeauthor{bagrow2025softly}~\cite{bagrow2025softly}, we follow the minimum description length principle and seek models that minimize the total number of bits required to encode the model and the data given the model:
\begin{equation}
\mathcal{L}_\text{MDL} = \mathcal{L}_\text{model} + \mathcal{L}_{\text{model}|\text{data}}.
\end{equation}
For data loss, we take the MSE, 
$\mathcal{L}_{\text{model}|\text{data}}\big(f(\vec{x}_i; \bm{\theta}), \vec{y}_i\big) = (1/n)\sum_{i=1}^{n} \| f(\vec{x}_i; \bm{\theta}) - \vec{y}_i \|^2$, which is proportional to the negative log-likelihood of the data given the model for iid normal errors.
And for model description length, making a BIC-style approximation~\cite{schwarz1978estimating}, we have
\begin{equation}
\mathcal{L}_\text{model} = \frac{\log n}{n} \lznorm{\bm\theta},
\end{equation}
with the $\lznorm{\bm\theta} \approx \mathcal{L}_C$ term capturing model complexity based on the number of open gates.

For model complexity, first consider the case of no xgates. 
The complexity is given by
\begin{equation}
\mathcal{L}_C =
\sum_{\ell=0}^{L-1} \sum_{j} \mathbb{E}[z_{\ell+1,j}] \left( c_{\ell+1,j} + \sum_{i} \mathbb{E}[z_{\ell ij}] \, c_{\ell ij} \right),
\label{eqn:approx-model-complexity}
\end{equation}
where the $\{\mathbb{E}[z]\}$ are given by Eq.~\eqref{eqn:expected-z} and we take $z_{Lj} =1$ (no ngates on outputs).
Equation~\eqref{eqn:approx-model-complexity} introduces tunable complexity costs:
$c_{\ell j}$ is the complexity cost for node $j$ in layer $\ell$, and $c_{\ell ij}$ is the complexity cost for the edge from node $i$ to node $j$ in layer $\ell$.
For simplicity, in this work we fix $c=1$ for all nodes and edges, but it may be useful to use $\{c\}$ to weight different parts of the model with different complexities accordingly.
Define the per-layer complexity $C_\ell$ as the inner sum of Eq.~\eqref{eqn:approx-model-complexity}:
$\sum_{j} \mathbb{E}[z_{\ell+1,j}] \left( c_{\ell+1,j} + \sum_{i} \mathbb{E}[z_{\ell ij}] \, c_{\ell ij} \right)  :=  C_\ell$,
so that $\mathcal{L}_C = \sum_{\ell=0}^{L-1} C_\ell$.

With xgates, the complexity of the inference path is the complexity of the exit head used plus the complexity of all trunk layers preceding it.
Exit head $k$ is used with probability $\pi_k$. 
The probability of using trunk layer $\ell$ is the probability of using an exit $k > \ell$, which is $\sum_{k > \ell} \pi_k$ since the exits are mutually exclusive.
Combining this with the per-layer complexities, and noting $\sum_{i,j} \mathbb{E}[z^{\text{h}}_{\ell ij}] \, c^{\text{h}}_{\ell ij}  := C_\ell^{\text{h}}$ as the complexity of exit head $\ell$
(exit heads have egates $z^\text{h}$ but not ngates),
we have
\begin{equation}
\mathcal{L}_C =
    \sum_{\ell=0}^{L-2}
    \left(\sum_{k>\ell} \pi_k\right) C_\ell
  \;+\;
    \sum_{\ell=0}^{L-1} \pi_\ell \, C_\ell^{\text{h}}
  \;+\;
    c^\text{x}(L - 1),
\label{eqn:reachability-mdl-compact}
\end{equation}
where the final term is the cost of the exit gate's $L{-}1$ free parameters $\{\pi_\ell\}$.

The weighting by exit probability in Eq.~\eqref{eqn:reachability-mdl-compact} arises from MDL principles. 
The description length of path $k$ (using exit $k$) is the cost of trunk layers $0, \ldots, k{-}1$ plus exit head $k$:
\begin{equation}
  \mathrm{DL}(k) = \sum_{\ell=0}^{k-1} C_\ell + C_k^{\text{h}}.
\end{equation}
The expected description length under the exit distribution is
\begin{align}
  \mathbb{E}_\pi[\mathrm{DL}]
  &= \sum_{k=0}^{L-1} \pi_k \left[\sum_{\ell=0}^{k-1} C_\ell + C_k^{\text{h}}\right] \\
  &= \sum_{k=0}^{L-1} \pi_k \sum_{\ell=0}^{k-1} C_\ell
    \;+\; \sum_{k=0}^{L-1} \pi_k \, C_k^{\text{h}}.
\end{align}
Swapping the order of summation in the trunk term,
\begin{equation}
  \sum_{k=0}^{L-1} \pi_k \sum_{\ell=0}^{k-1} C_\ell
  = \sum_{\ell=0}^{L-2} C_\ell \sum_{k=\ell+1}^{L-1} \pi_k
  = \sum_{\ell=0}^{L-2} \left(\sum_{k>\ell} \pi_k\right) C_\ell,
\end{equation}
recovers the first term of Eq.~\eqref{eqn:reachability-mdl-compact}.
The rewritten $\mathbb{E}_\pi[\mathrm{DL}]$ plus the exit gate cost $c^{\text{x}}(L{-}1)$ is the full complexity penalty.

\subsection{Training and inference}
\label{subsec:training-inference}

With our loss function specified, training can proceed.
We differentiably sample $\{z_{\ell ij}\} \cup \{z_{\ell j}\}$ per Sec.~\ref{subsec:background:diff-l0-sparse} and sample $\bm{\eta}$ per Sec.~\ref{subsec:background:gumbel-softmax}.
The data loss uses the stochastic Gumbel-Softmax samples: $\mathcal{L}_{\text{data}} = \sum_{k=0}^{L-1} \eta_k \, \text{MSE}_k$, where $\text{MSE}_k$ is the mean squared error at exit $k$, while the complexity term (Eq.~\eqref{eqn:reachability-mdl-compact}) uses the categorical parameters $\pi_k$ directly, mirroring the egates, where training uses stochastic samples but the regularizer uses $\mathbb{E}[\tilde{z}]$.

Training proceeds in phases. 
First, the trunk trains alone (warmup), allowing spline activations to stabilize before egates and FCs become active. 
FCs are disabled initially to prevent them from dominating before trunk features develop.
Exits are introduced at the start with xgate temperature $\tau_x$ annealed from high (near-uniform, providing deep supervision to all exits) to low (near-one-hot, committing to a single exit). 
Specific schedules and parameter values are given in App.~\ref{sec:methods}.

Standard practice with spline-based KANs is to periodically update their spline grids so knots better cover the input ranges of activation functions as they shift during training.
This shifting happens particularly at warmup boundaries when FCs or egates come online, so we use bursts of grid updates triggered at each warmup boundary.
See App.~\ref{sec:methods} for scheduling details.

At inference, we deterministically threshold egates and ngates, $\hat{z} = \mathbb{I}[\mathbb{E}[\tilde{z}] > 1/2]$, following \citeauthor{bagrow2025softly}~\cite{bagrow2025softly}. 
Unlike the estimator of \citeauthor{louizos2018learning}~\cite{louizos2018learning} (Eq.~\eqref{eqn:louizos-test-estimator}), this guarantees binary gate values, yielding a fully discrete architecture. 
The exit is selected as $\hat{k} = \arg\max_k \pi_k$.

We discuss further setup and training details in Methods (App.~\ref{sec:methods}).

\section{Results}
\label{sec:results}

To study the effects of sparsification and deep supervision, we adopt a $2 \times 2 \times 2$ factorial experiment, comparing KANs with and without egates, forward connections, and xgates.
Node gates (ngates) are a special case of egates, meaning edge-level sparsity is the more general mechanism, so we leave ngates to future work.
We refer to conditions using `E,' `F,' and `X' such that `E' is egate-only, `FX' is forward connections plus xgates (no egates), and so forth. 
Conditions are implemented as ablations, as removing FCs, setting $\alpha \to \infty$ or setting $\bm{\pi} = [0, \ldots, 0, 1]$ can each be applied independently; applying all three recovers a standard KAN, our baseline.
These ablations allow controlled comparison of the contributions of sparsification and deep supervision.

We report relative performance values in this section, comparing each condition's accuracy and complexity against the baseline; see Appendix~\ref{sec:absolute-results} for absolute values.
To measure accuracy, we compute the test RMSE for each condition, averaged over 10 independent seeds.
Likewise, we use two measures of complexity: the number of contributing edges and the degree of nesting or compositional depth (see Methods, App.~\ref{sec:methods}). 
Results at a given $\beta$ are not directly comparable across conditions, as the effective regularization differs, so we compute the Pareto hypervolume (HV)~\cite{zitzler2003performance,ishibuchi2018specify}, a scalar summary of how well a condition trades off accuracy against complexity across the full $\beta$ sweep.
See App.~\ref{sec:methods} for details of complexity and HV computation as well as specific details of architectures, other hyperparameters, and training.

\subsection{Function approximation}
\label{subsec:results:function-approximation}

We begin fitting different conditions to data from an example function, $z = \sin\left(x + y^2\right)$, shown in Fig.~\ref{fig:anecdote}.
While simple, this function tests for composition, as the network must learn to compose $\sin$ and $(\cdot)^2$.
The baseline condition (panel A) fits the data fairly well but none of the individual activations resemble the underlying functional components; the KAN lacks symbolic fidelity~\cite{bagrow2025softly}.
Adding egates (B) improves the fit and the first layer of activation functions now resemble $x$ and $y^2$, but the $\sin(\cdot)$ activation is spread over the remaining two layers.
Combining egates with either FCs (C) or exits (D) gives the network a mechanism to select the appropriate compositional depth and number of activation functions.

\begin{figure}[t]
    \centering
    \includegraphics[width=0.475\columnwidth]{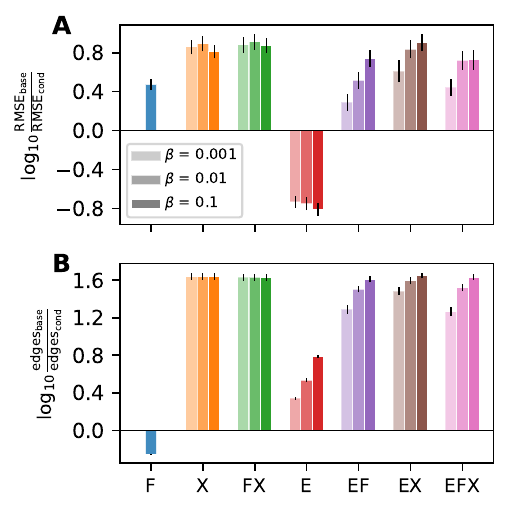}
    \caption{Model accuracy and size relative to KAN baseline. Log-ratios of (\textbf{A}) test RMSE  and (\textbf{B}) contributing edges, pooled across Nguyen problems F1--F10 (see also App.~\ref{sec:absolute-results}) and broken down by condition and $\beta$. 
    Positive values indicate conditions have lower RMSE or fewer contributing edges than baseline.
    Errorbars denote $\pm 1$ SEM on log-ratios.}
    \label{fig:logratio-nguyen}
\end{figure}

Next, Fig.~\ref{fig:logratio-nguyen} shows results on the first ten problems (8 univariate, 2 bivariate) of the Nguyen symbolic regression benchmark~\cite{uy2011semantically}.
(See Methods, App.~\ref{sec:methods} for details, Fig.~\ref{fig:logratio-nguyen-perproblem} for per-problem breakdown, and Tables~\ref{tab:raw-F01-raw-F02}--\ref{tab:raw-F09-raw-F10} for exact values.)
KANs often found highly accurate networks, always with fewer edges than baseline (except F, which can only add edges).
One notable exception is condition E (egate-only), which found smaller networks than baseline but at the cost of accuracy. 
Conditions featuring a depth selection mechanism (i.e., containing at least X or at least both E and F) were simultaneously smaller and more accurate than baseline.
While Fig.~\ref{fig:logratio-nguyen} is pooled over the ten problems, the interpretation holds across individual problems (Fig.~\ref{fig:logratio-nguyen-perproblem}, Tables~\ref{tab:raw-F01-raw-F02}--\ref{tab:raw-F09-raw-F10}): all conditions resulted in models smaller than baseline (except F), E led to worse accuracy than baseline on all problems, and other conditions generally led to both smaller and more accurate models.

\subsection{Forecasting dynamical systems}
\label{subsec:results:dynamical-systems}

KANs have shown success modeling dynamical systems~\cite{koenig2024kan,PhysRevResearch.7.023037,Bagrow_2025}, so we conducted our $2 \times 2 \times 2$ experiment on two exemplar dynamical systems.
The first is the \textit{Ikeda map}~\cite{ikeda1979multiple,hammel1985global}:
\begin{equation}
\begin{gathered}
x_{n+1} = 1 + \mu \left(x_n \cos\left(\phi_n\right)- y_n \sin\left(\phi_n\right)\right),\\
y_{n+1} = \mu\left(x_n \sin(\phi_n) + y_n \cos(\phi_n) \right),
\label{eqn:ikeda}
\end{gathered}
\end{equation}
where $\phi_n = 0.4 - 6\left(1 + x_n^2 + y_n^2\right)^{-1}$ and bifurcation parameter $\mu = 0.9$.
KANs are well-suited to modeling the Ikeda map due to its compositional nature whereas sparse regression methods such as SINDy~\cite{sindy2016} struggle~\cite{PhysRevResearch.7.023037}.

The second system is a continuous-time \textit{three-species ecosystem}:
\begin{equation}
\begin{gathered}
\frac{dN}{dt} = N\left(1 - \frac{N}{K}\right) - x_p y_p \frac{NP}{N + N_0}, \\
\frac{dP}{dt} = x_p P\left(y_p \frac{N}{N + N_0} - 1\right) - x_q y_q \frac{PQ}{P + P_0}, \\
\frac{dQ}{dt} = x_q Q\left(y_q \frac{P}{P + P_0} - 1\right),
\end{gathered}
\label{eqn:food}
\end{equation}
where $N$, $P$, and $Q$ denote the populations of primary producers, herbivores, and carnivores, respectively, with carrying capacity $K$ acting as the bifurcation parameter.
Following~\cite{mccann1994nonlinear}, we use $K = 0.98$, $x_p = 0.4$, $y_p = 2.009$, $x_q = 0.08$, $y_q = 2.876$, $N_0 = 0.16129$, and $P_0 = 0.5$, which produces chaotic dynamics.
Data for both systems were generated and partitioned into training and testing sets following~\citeauthor{PhysRevResearch.7.023037}~\cite{PhysRevResearch.7.023037}.
We used architectures of $[2, 4, 4, 4, 2]$ for the Ikeda map and $[3, 3, 3, 3]$ for the ecosystem, consistent with prior work~\cite{Bagrow_2025,bagrow2025softly}.

\begin{figure*}[t]
    \centering
    \includegraphics[width=0.9\textwidth]{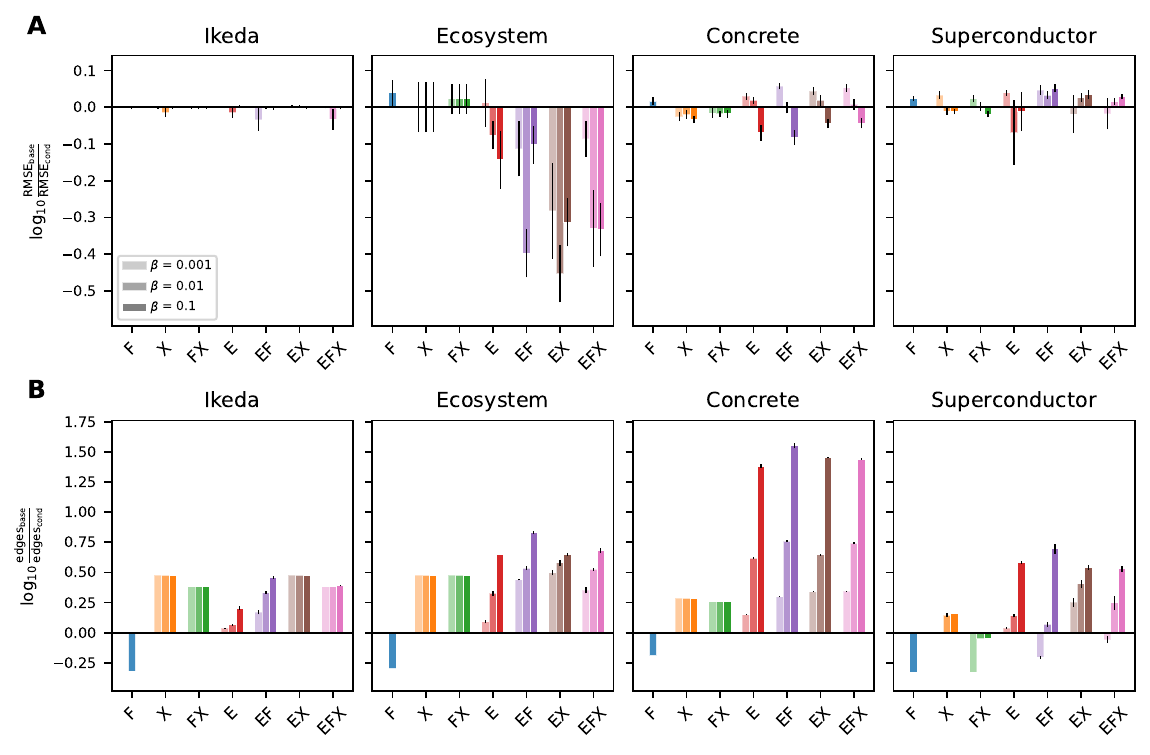}
    \caption{Model accuracy and size relative to KAN baseline. Log-ratios of (\textbf{A}) test RMSE and (\textbf{B}) contributing edges for dynamical systems models (Ikeda, Ecosystem) and real-world datasets (Concrete, Superconductor), broken down by condition and $\beta$.
    Positive values indicate lower RMSE or fewer contributing edges.
    Errorbars denote $\pm 1$ SEM on log-ratios.}
    \label{fig:logratio-sweepall-perdata}
\end{figure*}

The results for both systems are shown in the first two columns of Fig.~\ref{fig:logratio-sweepall-perdata}.
For the Ikeda map, all conditions that can achieve smaller models than baseline do, and generally with little to no loss in accuracy (multi-step test RMSE relative to baseline). 
The reduction in model size can be substantial: for EX, for instance, models used 16 edges compared to 48 for baseline (App.~\ref{sec:absolute-results}), without loss in accuracy.
For the ecosystem model, however, the results are more mixed.
Models are again smaller than baseline, but conditions with egates did so at the cost of accuracy, sometimes substantially so.
This system in particular appears prone to overregularization.
Consistent with prior findings~\cite{bagrow2025softly}, both systems exhibited self-sparsification, where egates close even without explicit regularization ($\beta = 0$, App.~\ref{sec:absolute-results}, Table~\ref{tab:raw-ikeda-raw-ecosystem}).

\subsection{Real-world data}
\label{subsec:results:real-world-data}

We conducted our $2 \times 2 \times 2$ experiment on two real-world datasets:

\textit{Compressive strength of concrete}---%
The task is to predict compressive strength (in MPa) of concrete from sample properties.
The dataset comprises 1030 samples with eight features:
cement, blast furnace slag, fly ash, water, superplasticizer, coarse aggregate, fine aggregate (all measured in kg/m$^{3}$), and age in days.
Since compressive strength is known to depend on the water-to-cement ratio~\cite{neville2011properties}, we included this as a derived feature.
We also included total binder (cement plus slag and fly ash), total aggregate (coarse plus fine), and the water-to-binder ratio, which extends the water-to-cement relationship to account for supplementary cementitious materials.
Because strength gain follows an approximately logarithmic relationship with curing time~\cite{neville2011properties}, we additionally include $\log(\text{age}+1)$ as a transformed age variable. 
An 80/20 train/test split was used for modeling.
Data were collected by \citeauthor{YEH19981797}~\cite{YEH19981797,YEHconcrete2}.

\textit{Critical temperature of superconductors}---%
The task is to predict the critical temperature $T_c$ (in K) of superconductors from their material properties.
The original dataset contains numerous features and derived statistics; following prior KAN modeling~\cite{Bagrow_2025}, we focus on five representative features that capture composition, electronic structure, and bonding: number of elements, weighted mean valence, valence entropy, weighted mean first ionization energy, and mean electron affinity.
We drew 1000 samples each for training and testing, from Japan's National Institute for Materials Science superconductor database~\cite{hamidieh2018data,center-a}.

As shown in the last two columns of Fig.~\ref{fig:logratio-sweepall-perdata} (see also Table~\ref{tab:raw-concrete-raw-supercond}), architectures that enable depth selection, sparsification, or both led to much smaller KANs with little accuracy loss, and even sometimes improved accuracy, for both datasets.
For instance, EFX on Concrete at $\beta=0.01$ achieved 4.87 MPa RMSE compared to Baseline's 4.91 MPa, an improvement but only by a modest 0.81\%. 
However, it achieved that reduction with a model 18\% the size (64 contributing edges compared to 351).
In general, for Concrete we see modest improvements in accuracy at low-to-intermediate values of $\beta$, with noticeable reductions in size, while higher values of $\beta$ lead to very small models but slightly worse accuracy than baseline.
The results are much the same for Superconductor, although we see less extreme reductions in size compared to Concrete.

\subsection{Pareto analysis}

To compare systematically how conditions trade off accuracy and parsimony, we analyzed their Pareto fronts. 
Because different conditions have different regularization terms, we summarize each Pareto front with its hypervolume (HV), rather than comparing directly on $\beta$.
The HV is a scalar measure of how well the front covers the space of accuracy and size.
We compute the 3D HV of test RMSE (multi-step RMSE in the case of dynamical systems) against contributing edges and compositional depth, two measures of model size.
For details on the HV calculation, see Methods (App.~\ref{sec:methods}).

Table~\ref{tab:hv-3d} shows the HV for each condition on all $2 \times 2 \times 2$ experiments, normalized by the HV of the top-ranked condition (note that Nguyen is normalized before averaging over the ten problems).
The dominant condition on average is EX, but no one condition dominates on all problems, and EF and EFX also rank well.
E does poorly overall, and X and FX do well except for the real-world problems.

\begin{table}[t]
\centering
\small
\caption{Normalized hypervolume of Pareto front of RMSE vs.\ contributing edges vs.\ compositional depth. Higher is better. Boldface marks best within each column. Dashes indicate ungated conditions (no $\beta$ sweep). 
Nguyen values are normalized per problem then averaged.}
\label{tab:hv-3d}
\setlength{\tabcolsep}{5pt}
\begin{tabular}{l c c c c c c}
\toprule
Condition & Nguyen & Ikeda & Ecosystem & Concrete & Supercond. & Mean \\
\midrule
Baseline & --- & --- & --- & --- & --- & --- \\
F & --- & --- & --- & --- & --- & --- \\
X & \textbf{0.999} & 0.996 & 0.908 & 0.396 & 0.292 & 0.718 \\
FX & 0.998 & 0.952 & 0.934 & 0.383 & 0.215 & 0.696 \\
E & 0.106 & 0.142 & 0.123 & 0.134 & 0.115 & 0.124 \\
EF & 0.993 & 0.994 & \textbf{1.000} & 0.935 & 0.702 & 0.925 \\
EX & 0.996 & \textbf{1.000} & 0.910 & 0.999 & \textbf{1.000} & \textbf{0.981} \\
EFX & \textbf{0.999} & 0.956 & 0.946 & \textbf{1.000} & 0.845 & 0.949 \\
\bottomrule
\end{tabular}
\end{table}

\section{Discussion}
\label{sec:discussion}

Our experiments provide the following insights:
Egates are necessary but not sufficient, as E typically harms accuracy. 
The best performing conditions are generally EF, EX, or EFX. 
This underscores the importance of depth selection, as E alone can reduce the number of edges but not the compositional depth.
Either F or X provides depth selection, as well as deep supervision during training.
Indeed, EX ranked highest on average HV but without one dominant condition, a practitioner might consider using EFX, the maximally expressive architecture.
This is, in some sense, surprising: FCs and exit heads could interfere with each other, as both provide alternative pathways to the output, yet we see they are more complementary than redundant.

While many conditions improve parsimony without sacrificing accuracy, and sometimes even improve it, the architectures still require hyperparameter tuning, particularly of $\beta$ but also architecture shape (depth and width), learning rate, and initialization values for gates and activation functions~\cite{rigas2026initialization}.
This is a standard hyperparameter search, a common practical concern in ML. 
Tools such as Optuna~\cite{10.1145/3292500.3330701} may be useful for fitting overprovisioned KANs. 
Our results here do not exhaustively explore the hyperparameter space and there may be unrealized gains.

Beyond the cost of increased hyperparameters, 
we also note that our architectural mechanisms can be sensitive to initialization values and seed.
Additionally, the differentiable gate used for edges is known to have variance issues during training~\cite{gale2019state,hoefler2021sparsity}.
The ecosystem model was prone to overregularization, particularly at high $\beta$, suggesting that sparsification requires sufficient architectural slack to be effective.
Our study also did not explore node gates because they are a special case of edge gates. 
However, node gates can provide group or structured sparsity, which may prove useful for accelerating the learning of some deep KANs.

Our study suggests some fruitful avenues for future work.
Instead of pruning an overprovisioned KAN, it may be more efficient to grow an underprovisioned KAN, an approach we have not explored.
Likewise, besides exploring structured sparsity with node gates, adding input gates to the KAN to allow for variable selection may be worthwhile.
We also did not consider the effects of complexity weights ($c_{\ell ij}$) to reflect computational cost or structural preferences; while their flexibility may complicate matters, making them most useful in domain-specific applications, they may yield increased interpretability.
Finally, learning with overprovisioned KANs should be studied in the context of symbolic regression~\cite{bagrow2025softly}, as symbolic regression is one of the KAN's most unique and promising applications~\cite{liu2025kan}.

In summary, to best fit a KAN without knowing \textit{a priori} the optimal architecture, we suggest using edge gates in combination with an overprovisioned, deeply supervised architecture capable of providing depth selection.
Forward connections provide implicit depth selection but only in concert with edge gates, while multi-exits offer explicit depth selection. 
Both provide deep supervision.
Overprovisioning and sparsification were synergistic, with the combination typically outperforming either component alone.
Crucially, all three architectural components are differentiable, allowing for gradient-based optimization of the KAN architecture during training, and making the full EFX architecture a safe default.
Differentiable sparsification transforms architecture search from a discrete hyperparameter choice into a learnable component of training, requiring only standard gradient-based optimization.
More broadly, these results offer a principled mechanism to effectively incorporate future advancements in deep learning into the interpretable domain of scientific machine learning.

\bibliographystyle{unsrtnat}
\bibliography{references}

\appendix

\section{Methods}
\label{sec:methods}

\subsection{Evaluation measures}
\label{subsec:methods:evaluation-measures}

\paragraph{Contributing edges}
After a model is trained, we evaluate its predictive performance on held-out test data, but we are also interested in its complexity, as more parsimonious models will be preferable given they are still accurate.
Our loss function is a training signal of complexity, but we also need to measure the realized complexity of the model, which we can do with the number of activation functions that remain in the model.
However, for gated models, it may be possible due to incomplete convergence for activation functions to have open gates but not contribute to the output of the model.
Active but disconnected edges may be present in the model.
We therefore define \emph{contributing edges} as those activation functions that actually contribute to the output of the model.
However, there is a subtlety: there may be contributing edges that do not sit on a path from input to output.
They contribute to the model, but only as bias terms.
One may even encounter multiple edges collectively serving as a single bias term, and if this occurs we should merge them into a single contributing edge when measuring model complexity.
In our experience, these issues are rare, but nevertheless it is worth accounting for the small number of such edges when measuring complexity.

With these considerations in mind, to measure the complexity of trained models with egates and/or xgates we use a two-step backward--forward trace.
First, mask out all inactive edges (closed egate, ngate, and/or xgate).
Then trace backward from the output node(s) to the input(s) to find all connected edges.
Next, trace forward from the inputs along the connected edges only, finding all paths that connect input to output.
(The maximum length of these paths is the \emph{compositional depth} of the model.)
Any connected edges not on these paths are the bias terms.
Remove the paths and count the number of connected components that remain to get the number of independent bias terms.
Lastly, the number of contributing edges is the number of connected edges in the paths plus the number of connected components off the paths.

\paragraph{Pareto hypervolume}
To compare conditions across all values of $\beta$, we compute the Pareto hypervolume (HV)~\cite{zitzler2003performance} of each condition's Pareto front in the space of RMSE, contributing edges, and compositional depth (all minimized). 
Points are seed-medians at each $\beta$. 
The reference point is set to $1.1$ times the per-objective maximum across all conditions~\cite{ishibuchi2018specify}. 
Values are normalized by the maximum HV across conditions per dataset. 
Nguyen HVs are computed per-problem then averaged.

\subsection{Experimental details}
\label{subsec:methods:experimental-details}

KAN activations used B-splines with $G=10$ grid intervals and degree $K=3$ for all experiments; grid refinement was not used.
Spline grids and coefficients were initialized as described in~\citeauthor{liu2025kan}~\cite{liu2025kan}.
For spline grid updates~\cite{liu2025kan}, knots were re-positioned 10 times at intervals of 5 epochs, triggered at training start, FC activation (FC warmup), and $\beta$ onset (Warmup).
The SiLU base function (Sec.~\ref{subsec:background:kans}) was used for all experiments.

For egates, we used temperature $\tau = 2/3$ and stretch parameters $\gamma = -0.1$, $\zeta = 1.1$.
Unless otherwise noted, egates were initialized with $\alpha = -1$.
For conditions without egates (Baseline, F, X, FX), egate logits were fixed at $\alpha = 20$ with gradients disabled and $\beta = 0$ throughout. 
For xgates, temperature annealing  $\tau_x = 5.0 \to 0.1$ with exponential decay was used.
For conditions without F, FCs were removed; for conditions without X, exit heads were removed from the network prior to training.

Training used Adam~\cite{kingma2014adam} with default parameters (no weight decay) and a constant learning rate of $10^{-3}$ for both coefficients and gates.
For conditions with egates, no complexity pressure was exerted during warmup; egates were present to allow stochastic sampling but $\beta = 0$ until warmup ended. 
Forward connections, if present, were disabled for the first 100 epochs (FC warmup). 
Early stopping was not used.
Dataset-specific hyperparameters are summarized in Table~\ref{tab:hyperparams}.

\begin{table}%
\centering
\small
\caption{Hyperparameters for the $2\times 2\times 2$ factorial experiments.}
\label{tab:hyperparams}
\setlength{\tabcolsep}{4pt}
\begin{tabular}{l c c c c c}
\toprule
 & Nguyen & Ikeda & Ecosystem & Concrete & Superconductor \\
\midrule
Architecture        & $[n, 5, 5, 5, 1]$ & $[2, 4, 4, 4, 2]$ & $[3, 3, 3, 3]$ & $[13, 13, 13, 1]$ & $[5, 5, 5, 5, 1]$ \\
Epochs              & 10\,000 & 4\,000 & 10\,000 & 5\,000 & 5\,000 \\
Batch size          & 128 & 128 & 128 & 64 & 64 \\
$\beta$ values      & 4 values$^a$ & \multicolumn{2}{c}{6 values$^b$} & \multicolumn{2}{c}{7 values$^c$} \\
Warmup (epochs)     & 200 & 200 & 200 & 500 & 500 \\
FC warmup (epochs)  & 100 & 100 & 100 & 100 & 100 \\
Gate init $\alpha$  & $-1$ & $-2$ & $-2$ & $-1$ & $-1$ \\
Grid updates        & yes & no & no & yes & yes \\
$n_\text{train}$    & 1\,024 & 6\,400 & 3\,000 & 824 & 1\,000 \\
$n_\text{test}$     & 256 & 1\,600 & 750 & 206 & 1\,000 \\
\bottomrule
\end{tabular}

\vspace{2pt}
{\footnotesize
$^a$\,$\beta_a = \{0.0, 0.001, 0.01, 0.1\}$\quad
$^b$\,$\beta_b = \beta_a \cup \{0.005, 0.05\}$\quad
$^c$\,$\beta_c = \beta_b \cup \{0.5\}$
}
\end{table}

KAN networks were implemented in PyTorch v2.8.0.
Code will be made available upon publication.

\textit{Function approximation (Sec.~\ref{subsec:results:function-approximation})}---
For Fig.\ref{fig:anecdote},
we generated 1024 training and 256 test points on $x, y \in [-2, 2]$, with target $z = \sin(x +y^{2})$.
The model used architecture $[2, 2, 1, 1]$ and trained for 3000 epochs with batch size 64 and $\beta = 0.2$ with a 200-epoch warmup (100-epoch FC warmup).
For the Nguyen benchmark, we evaluated the first 10 Nguyen problems~\cite{uy2011semantically} following the parameters in Table~\ref{tab:hyperparams}.
The problems and data domains used in this work are:
F1: $y = x^3 + x^2 + x$;
F2: $y = x^4 + x^3 + x^2 + x$;
F3: $y = x^5 + x^4 + x^3 + x^2 + x$;
F4: $y = x^6 + x^5 + x^4 + x^3 + x^2 + x$;
F5: $y = \sin(x^2)\cos(x) - 1$;
F6: $y = \sin(x) + \sin(x + x^2)$;
F7: $y = \log(x+1) + \log(x^2+1)$;
F8: $y = \sqrt{x}$.
F1--F8 are univariate with $x \in [-1, 1]$ except F7 ($x \in [0, 2]$) and F8 ($x \in [0, 4]$).
F9: $z = \sin(x) + \sin(y^2)$, $x, y \in [-1, 1]$;
F10: $z = 2\sin(x)\cos(y)$, $x, y \in [-\pi, \pi]$.
\textit{Dynamical systems (Sec.~\ref{subsec:results:dynamical-systems})}---
The data generating process and parameter values are given in Sec.~\ref{subsec:results:dynamical-systems} and follow \citeauthor{PhysRevResearch.7.023037}~\cite{PhysRevResearch.7.023037}.
\textit{Real-world data (Sec.~\ref{subsec:results:real-world-data})}---
For the concrete compressive strength prediction task, we used the UCI concrete dataset~\cite{YEH19981797,YEHconcrete2} with 8 raw features augmented by 5 derived features (water-cement ratio, water-binder ratio, total binder, total aggregate, log age), for 13 total.
For superconductor critical temperature prediction, we used 5 features from the UCI superconductor dataset~\cite{hamidieh2018data,center-a}: number of elements, weighted mean valence, weighted mean first ionization energy, mean electron affinity, and valence entropy.

\section{Supplementary results}
\label{sec:absolute-results}

\begin{figure}
    \centering
    \includegraphics[width=0.9\textwidth]{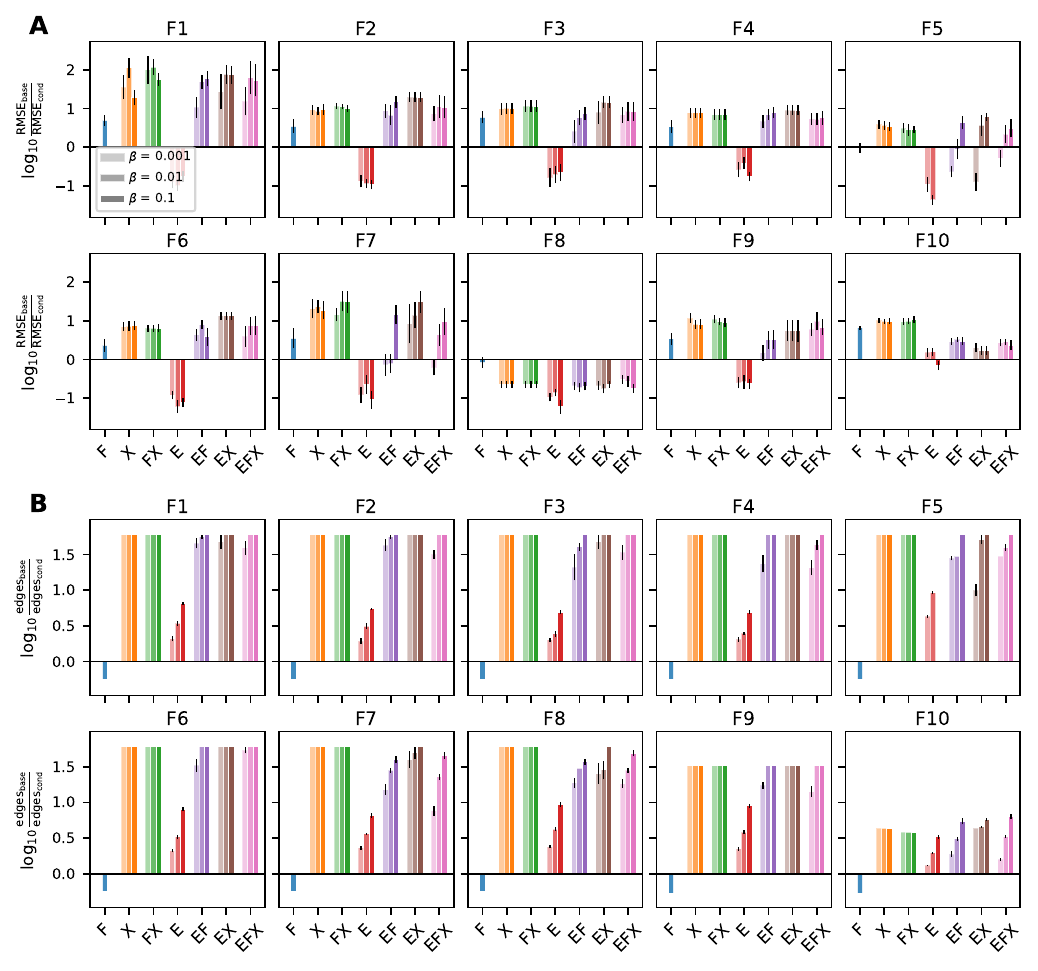}
    \caption{Log-ratio of RMSE (A) and contributing edges (B) relative to Baseline for each Nguyen problem (F1--F10).}
    \label{fig:logratio-nguyen-perproblem}
\end{figure}

Complementing the pooled Nguyen relative results (Fig.~\ref{fig:logratio-nguyen}), Fig.~\ref{fig:logratio-nguyen-perproblem} shows the relative results per problem. 
We also provide absolute values for all analyzed models:
Tables~\ref{tab:raw-ikeda-raw-ecosystem} and~\ref{tab:raw-concrete-raw-supercond} report RMSE, contributing edges, and compositional depth (median [min, max] across 10 seeds) for the dynamical systems and real-world datasets, respectively.
Tables~\ref{tab:raw-F01-raw-F02}--\ref{tab:raw-F09-raw-F10} report the same for each Nguyen problem.

\begin{table}[p]
\centering
\small
\caption{Absolute results: Ikeda (left) and Ecosystem (right). Median [min, max] across 10 seeds.}
\label{tab:raw-ikeda-raw-ecosystem}
\footnotesize
\begin{minipage}[t]{0.49\textwidth}
\centering
\setlength{\tabcolsep}{3pt}
\begin{tabular}{l r r r r}
\toprule
\multicolumn{5}{c}{\textbf{Ikeda}} \\
\midrule
Cond. & $\beta$ & RMSE & Cont.\ Edges & Depth \\
\midrule
Baseline & --- & 0.8622 [0.8365, 0.8799] & 48 [48, 48] & 4 [4, 4] \\
\addlinespace[2pt]
F & --- & 0.8663 [0.8567, 0.8789] & 100 [100, 100] & 4 [4, 4] \\
\addlinespace[2pt]
X & 0.0 & \textit{0.8589} [0.8458, 0.8700] & 48 [32, 48] & 4 [3, 4] \\
 & 0.001 & \textit{0.8616} [0.8504, 0.8855] & 16 [16, 16] & 2 [2, 2] \\
 & 0.005 & \textit{0.8621} [0.8521, 0.8720] & 16 [16, 16] & 2 [2, 2] \\
 & 0.01 & 0.8654 [0.8337, 1.1472] & 16 [16, 16] & 2 [2, 2] \\
 & 0.05 & 0.8636 [0.8555, 0.8839] & 16 [16, 16] & 2 [2, 2] \\
 & 0.1 & \textit{0.8614} [0.8415, 0.8777] & 16 [16, 16] & 2 [2, 2] \\
\addlinespace[2pt]
FX & 0.0 & 0.8628 [0.8458, 0.8759] & 52 [52, 100] & 3 [3, 4] \\
 & 0.001 & 0.8636 [0.8467, 0.8797] & 20 [20, 20] & 2 [2, 2] \\
 & 0.005 & 0.8627 [0.8490, 0.8775] & 20 [20, 20] & 2 [2, 2] \\
 & 0.01 & 0.8673 [0.8505, 0.8808] & 20 [20, 20] & 2 [2, 2] \\
 & 0.05 & 0.8647 [0.8412, 0.8822] & 20 [20, 20] & 2 [2, 2] \\
 & 0.1 & 0.8631 [0.8477, 0.8823] & 20 [20, 20] & 2 [2, 2] \\
\addlinespace[2pt]
E & 0.0 & \textit{0.8587} [0.8507, 0.8834] & 45 [42, 48] & 4 [4, 4] \\
 & 0.001 & \textit{0.8578} [0.8423, 0.8797] & 44 [42, 47] & 4 [4, 4] \\
 & 0.005 & 0.8632 [0.8499, 0.8922] & 43 [40, 46] & 4 [4, 4] \\
 & 0.01 & 0.8647 [0.8328, 1.2119] & 42 [39, 44] & 4 [4, 4] \\
 & 0.05 & 0.8640 [0.8512, 0.8889] & 34 [31, 39] & 4 [4, 4] \\
 & 0.1 & \textbf{0.8560} [0.8406, 0.8830] & 30 [25, 36] & 4 [4, 4] \\
\addlinespace[2pt]
EF & 0.0 & 0.8657 [0.8082, 0.8869] & 93 [85, 97] & 4 [4, 4] \\
 & 0.001 & 0.8625 [0.8306, 1.7936] & 32 [26, 40] & 3 [2, 3] \\
 & 0.005 & 0.8634 [0.7990, 0.8788] & 24 [23, 27] & 2 [2, 2] \\
 & 0.01 & 0.8652 [0.8473, 0.8808] & 23 [20, 24] & 2 [2, 2] \\
 & 0.05 & 0.8649 [0.8460, 0.8868] & 20 [16, 20] & 2 [2, 2] \\
 & 0.1 & 0.8678 [0.8492, 0.8939] & 16 [15, 19] & 2 [2, 2] \\
\addlinespace[2pt]
EX & 0.0 & \textit{0.8566} [0.8468, 0.8736] & 16 [16, 16] & 2 [2, 2] \\
 & 0.001 & \textit{0.8586} [0.8194, 0.8760] & 16 [16, 16] & 2 [2, 2] \\
 & 0.005 & 0.8689 [0.8401, 0.8906] & 16 [16, 16] & 2 [2, 2] \\
 & 0.01 & \textit{0.8568} [0.8350, 0.8731] & 16 [16, 16] & 2 [2, 2] \\
 & 0.05 & 0.8639 [0.8392, 0.8778] & 16 [16, 16] & 2 [2, 2] \\
 & 0.1 & \textit{0.8619} [0.8500, 0.8794] & 16 [16, 16] & 2 [2, 2] \\
\addlinespace[2pt]
EFX & 0.0 & \textit{0.8599} [0.8506, 0.9066] & 20 [20, 20] & 2 [2, 2] \\
 & 0.001 & \textit{0.8615} [0.8470, 0.8779] & 20 [20, 20] & 2 [2, 2] \\
 & 0.005 & 0.8667 [0.8453, 1.7395] & 20 [20, 20] & 2 [2, 2] \\
 & 0.01 & 0.8715 [0.8476, 1.7430] & 20 [20, 20] & 2 [2, 2] \\
 & 0.05 & 0.8653 [0.8490, 0.8913] & 20 [19, 20] & 2 [2, 2] \\
 & 0.1 & \textit{0.8580} [0.8466, 0.9027] & 20 [18, 20] & 2 [2, 2] \\
\bottomrule
\end{tabular}
\end{minipage}
~
\begin{minipage}[t]{0.49\textwidth}
\centering
\setlength{\tabcolsep}{3pt}
\begin{tabular}{l r r r r}
\toprule
\multicolumn{5}{c}{\textbf{Ecosystem}} \\
\midrule
Cond. & $\beta$ & RMSE & Cont.\ Edges & Depth \\
\midrule
Baseline & --- & 0.1717 [0.0726, 0.1957] & 27 [27, 27] & 3 [3, 3] \\
\addlinespace[2pt]
F & --- & \textit{0.1429} [0.1160, 0.1920] & 54 [54, 54] & 3 [3, 3] \\
\addlinespace[2pt]
X & 0.0 & \textit{0.1338} [0.0602, 0.2046] & 18 [18, 18] & 2 [2, 2] \\
 & 0.001 & 0.1903 [0.0762, 0.2404] & 9 [9, 9] & 1 [1, 1] \\
 & 0.005 & 0.1903 [0.0762, 0.2404] & 9 [9, 9] & 1 [1, 1] \\
 & 0.01 & 0.1903 [0.0762, 0.2404] & 9 [9, 9] & 1 [1, 1] \\
 & 0.05 & 0.1902 [0.0762, 0.2404] & 9 [9, 9] & 1 [1, 1] \\
 & 0.1 & 0.1903 [0.0762, 0.2404] & 9 [9, 9] & 1 [1, 1] \\
\addlinespace[2pt]
FX & 0.0 & \textbf{0.1013} [0.0649, 0.1842] & 27 [27, 27] & 2 [2, 2] \\
 & 0.001 & \textit{0.1549} [0.1126, 0.2026] & 9 [9, 9] & 1 [1, 1] \\
 & 0.005 & \textit{0.1550} [0.1126, 0.2026] & 9 [9, 9] & 1 [1, 1] \\
 & 0.01 & \textit{0.1549} [0.1126, 0.2026] & 9 [9, 9] & 1 [1, 1] \\
 & 0.05 & \textit{0.1549} [0.1126, 0.2026] & 9 [9, 9] & 1 [1, 1] \\
 & 0.1 & \textit{0.1549} [0.1126, 0.2026] & 9 [9, 9] & 1 [1, 1] \\
\addlinespace[2pt]
E & 0.0 & 0.1763 [0.0649, 0.2160] & 26 [23, 26] & 3 [3, 3] \\
 & 0.001 & 0.1785 [0.0529, 0.2213] & 22 [18, 25] & 3 [3, 3] \\
 & 0.005 & 0.1929 [0.1611, 0.2344] & 18 [13, 22] & 3 [3, 3] \\
 & 0.01 & 0.1866 [0.1509, 0.2955] & 13 [10, 16] & 3 [3, 3] \\
 & 0.05 & 0.1919 [0.1742, 0.2021] & 6 [6, 10] & 3 [3, 3] \\
 & 0.1 & 0.1899 [0.1751, 0.1974] & 6 [6, 6] & 3 [3, 3] \\
\addlinespace[2pt]
EF & 0.0 & \textit{0.1146} [0.0878, 0.2223] & 15 [10, 19] & 2 [1, 2] \\
 & 0.001 & 0.2009 [0.0840, 0.5105] & 10 [9, 10] & 1 [1, 1] \\
 & 0.005 & 0.4506 [0.2655, 0.5556] & 9 [8, 9] & 1 [1, 1] \\
 & 0.01 & 0.4611 [0.2526, 0.5795] & 8 [6, 9] & 1 [1, 1] \\
 & 0.05 & 0.2311 [0.1536, 0.4310] & 5 [4, 6] & 1 [1, 1] \\
 & 0.1 & 0.2025 [0.1620, 0.2557] & 4 [3, 5] & 1 [1, 1] \\
\addlinespace[2pt]
EX & 0.0 & 0.1794 [0.0978, 0.5881] & 11 [8, 17] & 2 [1, 2] \\
 & 0.001 & 0.3436 [0.0643, 1.0422] & 8 [7, 13] & 1 [1, 2] \\
 & 0.005 & 0.6588 [0.1999, 1.0440] & 8 [6, 9] & 1 [1, 1] \\
 & 0.01 & 0.5865 [0.2064, 0.7818] & 8 [5, 8] & 1 [1, 1] \\
 & 0.05 & 0.3102 [0.2069, 0.5578] & 7 [5, 7] & 1 [1, 1] \\
 & 0.1 & 0.3130 [0.2063, 0.6579] & 6 [5, 7] & 1 [1, 1] \\
\addlinespace[2pt]
EFX & 0.0 & \textit{0.1053} [0.0610, 0.2660] & 20 [19, 25] & 2 [2, 2] \\
 & 0.001 & 0.1996 [0.1114, 0.2885] & 12 [10, 17] & 2 [1, 2] \\
 & 0.005 & 0.2582 [0.1064, 0.5571] & 9 [8, 10] & 1 [1, 1] \\
 & 0.01 & 0.3475 [0.1036, 0.8208] & 8 [7, 9] & 1 [1, 1] \\
 & 0.05 & 0.3220 [0.2364, 0.4636] & 7 [5, 8] & 1 [1, 1] \\
 & 0.1 & 0.3180 [0.2090, 0.9322] & 6 [4, 7] & 1 [1, 1] \\
\bottomrule
\end{tabular}
\end{minipage}
\end{table}
\begin{table}[p]
\centering
\small
\caption{Absolute results: Concrete (left) and Superconductor (right). Median [min, max] across 10 seeds.}
\label{tab:raw-concrete-raw-supercond}
\footnotesize
\begin{minipage}[t]{0.49\textwidth}
\centering
\setlength{\tabcolsep}{3pt}
\begin{tabular}{l r r r r}
\toprule
\multicolumn{5}{c}{\textbf{Concrete}} \\
\midrule
Cond. & $\beta$ & RMSE (MPa) & Cont.\ Edges & Depth \\
\midrule
Baseline & --- & 4.91 [4.12, 6.56] & 351 [351, 351] & 3 [3, 3] \\
\addlinespace[2pt]
F & --- & \textit{4.82} [4.28, 5.92] & 546 [546, 546] & 3 [3, 3] \\
\addlinespace[2pt]
X & 0.0 & \textit{4.84} [4.33, 6.60] & 351 [351, 351] & 3 [3, 3] \\
 & 0.001 & 5.25 [4.81, 6.75] & 182 [182, 182] & 2 [2, 2] \\
 & 0.005 & 5.27 [4.73, 6.63] & 182 [182, 182] & 2 [2, 2] \\
 & 0.01 & 5.24 [4.78, 6.36] & 182 [182, 182] & 2 [2, 2] \\
 & 0.05 & 5.21 [4.61, 6.63] & 182 [182, 182] & 2 [2, 2] \\
 & 0.1 & 5.29 [4.95, 7.12] & 182 [182, 182] & 2 [2, 2] \\
 & 0.5 & 5.15 [4.85, 6.55] & 182 [182, 182] & 2 [2, 2] \\
\addlinespace[2pt]
FX & 0.0 & \textit{4.71} [4.03, 6.43] & 546 [546, 546] & 3 [3, 3] \\
 & 0.001 & 5.32 [4.45, 6.53] & 195 [195, 195] & 2 [2, 2] \\
 & 0.005 & 5.18 [4.36, 6.67] & 195 [195, 195] & 2 [2, 2] \\
 & 0.01 & 5.16 [4.39, 7.27] & 195 [195, 195] & 2 [2, 2] \\
 & 0.05 & 5.28 [4.40, 6.86] & 195 [195, 195] & 2 [2, 2] \\
 & 0.1 & 5.15 [4.28, 6.97] & 195 [195, 195] & 2 [2, 2] \\
 & 0.5 & 5.23 [4.32, 6.27] & 195 [195, 195] & 2 [2, 2] \\
\addlinespace[2pt]
E & 0.0 & \textit{4.35} [3.98, 5.99] & 351 [350, 351] & 3 [3, 3] \\
 & 0.001 & \textit{4.67} [4.12, 6.01] & 252 [238, 257] & 3 [3, 3] \\
 & 0.005 & \textit{4.65} [3.98, 5.88] & 120 [116, 138] & 3 [3, 3] \\
 & 0.01 & \textit{4.84} [4.20, 6.00] & 84 [77, 92] & 3 [3, 3] \\
 & 0.05 & 5.40 [4.48, 6.45] & 25 [18, 31] & 3 [3, 3] \\
 & 0.1 & 6.00 [4.82, 7.16] & 15 [12, 18] & 3 [3, 3] \\
 & 0.5 & 9.30 [7.26, 16.84] & 6 [4, 8] & 3 [3, 3] \\
\addlinespace[2pt]
EF & 0.0 & \textbf{4.12} [3.50, 5.80] & 546 [544, 546] & 3 [3, 3] \\
 & 0.001 & \textit{4.30} [4.00, 5.70] & 173 [165, 197] & 3 [3, 3] \\
 & 0.005 & \textit{4.72} [4.24, 5.77] & 92 [83, 102] & 3 [2, 3] \\
 & 0.01 & 5.19 [4.29, 6.04] & 60 [56, 67] & 2 [2, 2] \\
 & 0.05 & 5.05 [4.90, 6.93] & 20 [17, 24] & 2 [2, 2] \\
 & 0.1 & 6.30 [5.50, 6.90] & 10 [7, 12] & 2 [1, 2] \\
 & 0.5 & 7.29 [6.39, 8.43] & 4 [2, 5] & 1 [1, 2] \\
\addlinespace[2pt]
EX & 0.0 & \textit{4.35} [3.89, 5.62] & 182 [181, 182] & 2 [2, 2] \\
 & 0.001 & \textit{4.46} [3.94, 5.85] & 161 [154, 165] & 2 [2, 2] \\
 & 0.005 & \textit{4.84} [4.10, 5.74] & 108 [101, 113] & 2 [2, 2] \\
 & 0.01 & 4.93 [4.23, 5.70] & 79 [70, 89] & 2 [2, 2] \\
 & 0.05 & 5.56 [5.24, 6.64] & 13 [13, 13] & 1 [1, 1] \\
 & 0.1 & 5.56 [5.26, 6.64] & 12 [12, 13] & 1 [1, 1] \\
 & 0.5 & 6.10 [5.27, 6.82] & 8 [7, 9] & 1 [1, 1] \\
\addlinespace[2pt]
EFX & 0.0 & \textit{4.16} [3.87, 5.75] & 546 [542, 546] & 3 [3, 3] \\
 & 0.001 & \textit{4.36} [3.98, 6.03] & 161 [152, 165] & 2 [2, 2] \\
 & 0.005 & \textit{4.60} [4.11, 5.84] & 93 [83, 106] & 2 [2, 2] \\
 & 0.01 & \textit{4.87} [4.64, 5.79] & 64 [56, 70] & 2 [2, 2] \\
 & 0.05 & 5.30 [4.69, 6.64] & 19 [13, 24] & 2 [1, 2] \\
 & 0.1 & 5.57 [5.22, 6.66] & 13 [12, 13] & 1 [1, 1] \\
 & 0.5 & 5.91 [5.16, 7.06] & 9 [8, 10] & 1 [1, 1] \\
\bottomrule
\end{tabular}
\end{minipage}
~
\begin{minipage}[t]{0.49\textwidth}
\centering
\setlength{\tabcolsep}{3pt}
\begin{tabular}{l r r r r}
\toprule
\multicolumn{5}{c}{\textbf{Superconductor}} \\
\midrule
Cond. & $\beta$ & RMSE (K) & Cont.\ Edges & Depth \\
\midrule
Baseline & --- & 22.56 [19.92, 24.49] & 80 [80, 80] & 4 [4, 4] \\
\addlinespace[2pt]
F & --- & \textit{21.58} [19.50, 22.87] & 170 [170, 170] & 4 [4, 4] \\
\addlinespace[2pt]
X & 0.0 & \textit{21.77} [18.97, 22.50] & 80 [80, 80] & 4 [4, 4] \\
 & 0.001 & \textit{20.57} [19.38, 23.50] & 80 [80, 80] & 4 [4, 4] \\
 & 0.005 & \textit{22.01} [19.56, 26.00] & 80 [55, 80] & 4 [3, 4] \\
 & 0.01 & 22.82 [20.30, 25.77] & 55 [55, 80] & 3 [3, 4] \\
 & 0.05 & 22.96 [18.99, 24.10] & 55 [55, 80] & 3 [3, 4] \\
 & 0.1 & 22.94 [20.99, 24.67] & 55 [55, 55] & 3 [3, 3] \\
 & 0.5 & 22.78 [21.86, 24.30] & 55 [55, 80] & 3 [3, 4] \\
\addlinespace[2pt]
FX & 0.0 & \textit{21.09} [19.03, 22.80] & 170 [170, 170] & 4 [4, 4] \\
 & 0.001 & \textit{20.97} [18.94, 26.54] & 170 [170, 170] & 4 [4, 4] \\
 & 0.005 & 22.89 [21.53, 27.64] & 90 [90, 170] & 3 [3, 4] \\
 & 0.01 & \textit{22.27} [19.41, 25.56] & 90 [90, 90] & 3 [3, 3] \\
 & 0.05 & 22.67 [21.05, 27.42] & 90 [90, 90] & 3 [3, 3] \\
 & 0.1 & 22.70 [21.36, 26.70] & 90 [90, 90] & 3 [3, 3] \\
 & 0.5 & 22.82 [21.76, 27.73] & 90 [90, 90] & 3 [3, 3] \\
\addlinespace[2pt]
E & 0.0 & \textit{20.35} [18.88, 22.06] & 78 [66, 80] & 4 [4, 4] \\
 & 0.001 & \textit{20.06} [18.69, 21.93] & 76 [66, 79] & 4 [4, 4] \\
 & 0.005 & \textit{20.51} [18.05, 21.73] & 68 [54, 72] & 4 [4, 4] \\
 & 0.01 & \textit{21.19} [18.97, 174.79] & 60 [47, 67] & 4 [4, 4] \\
 & 0.05 & \textit{20.27} [17.81, 21.46] & 30 [28, 38] & 4 [4, 4] \\
 & 0.1 & \textit{20.60} [18.28, 72.25] & 21 [18, 24] & 4 [4, 4] \\
 & 0.5 & \textit{19.80} [18.29, 21.22] & 8 [5, 11] & 4 [4, 4] \\
\addlinespace[2pt]
EF & 0.0 & \textit{19.63} [17.96, 23.30] & 164 [136, 170] & 4 [4, 4] \\
 & 0.001 & \textit{19.66} [18.13, 25.63] & 132 [107, 149] & 4 [4, 4] \\
 & 0.005 & \textit{19.92} [18.74, 21.69] & 93 [78, 109] & 4 [4, 4] \\
 & 0.01 & \textit{20.92} [18.99, 22.79] & 70 [53, 85] & 4 [4, 4] \\
 & 0.05 & \textit{20.46} [20.08, 22.37] & 30 [18, 38] & 4 [2, 4] \\
 & 0.1 & \textbf{19.56} [17.68, 22.54] & 17 [8, 24] & 2 [2, 3] \\
 & 0.5 & \textit{21.06} [17.38, 26.06] & 5 [2, 9] & 2 [1, 2] \\
\addlinespace[2pt]
EX & 0.0 & \textit{20.58} [19.59, 22.78] & 53 [30, 55] & 3 [2, 3] \\
 & 0.001 & \textit{20.84} [19.53, 63.58] & 53 [29, 55] & 3 [2, 3] \\
 & 0.005 & \textit{20.54} [19.60, 22.48] & 30 [28, 55] & 2 [2, 3] \\
 & 0.01 & \textit{21.43} [18.31, 22.93] & 28 [25, 54] & 2 [2, 3] \\
 & 0.05 & \textit{20.30} [18.97, 23.34] & 24 [20, 28] & 2 [2, 2] \\
 & 0.1 & \textit{20.40} [18.48, 24.03] & 24 [18, 29] & 2 [2, 2] \\
 & 0.5 & \textit{20.99} [19.88, 27.58] & 3 [3, 4] & 1 [1, 1] \\
\addlinespace[2pt]
EFX & 0.0 & \textit{20.00} [18.09, 21.64] & 166 [86, 168] & 4 [3, 4] \\
 & 0.001 & \textit{20.99} [18.71, 56.63] & 84 [78, 147] & 3 [3, 4] \\
 & 0.005 & \textit{21.56} [19.84, 33.37] & 76 [73, 84] & 3 [3, 3] \\
 & 0.01 & \textit{21.40} [18.88, 24.11] & 33 [31, 78] & 2 [2, 3] \\
 & 0.05 & \textit{21.23} [20.06, 27.57] & 29 [25, 32] & 2 [2, 2] \\
 & 0.1 & \textit{21.24} [18.64, 23.32] & 24 [17, 31] & 2 [2, 2] \\
 & 0.5 & \textit{20.74} [18.41, 27.88] & 4 [3, 8] & 2 [1, 2] \\
\bottomrule
\end{tabular}
\end{minipage}
\end{table}

\begin{sidewaystable}[p]
\centering
\small
\caption{Absolute results: F1 (left) and F2 (right). Median [min, max] across 10 seeds.}
\label{tab:raw-F01-raw-F02}
\footnotesize
\begin{minipage}[t]{0.35\textwidth}
\centering
\setlength{\tabcolsep}{3pt}
\begin{tabular}{l r r r r}
\toprule
\multicolumn{5}{c}{\textbf{F1: $y = x^3 + x^2 + x$}} \\
\midrule
Cond. & $\beta$ & RMSE & Cont.\ Edges & Depth \\
\midrule
Baseline & --- & 0.0011 [0.0003, 0.0067] & 60 [60, 60] & 4 [4, 4] \\
\addlinespace[2pt]
F & --- & \textit{0.0002} [0.0001, 0.0004] & 106 [106, 106] & 4 [4, 4] \\
\addlinespace[2pt]
X & 0.0 & \textit{0.0003} [0.0000, 0.0021] & 10 [1, 10] & 2 [1, 2] \\
 & 0.001 & \textit{0.0000} [0.0000, 0.0004] & 1 [1, 1] & 1 [1, 1] \\
 & 0.01 & \textit{0.0000} [0.0000, 0.0003] & 1 [1, 1] & 1 [1, 1] \\
 & 0.1 & \textit{0.0001} [0.0000, 0.0003] & 1 [1, 1] & 1 [1, 1] \\
\addlinespace[2pt]
FX & 0.0 & \textit{0.0000} [0.0000, 0.0002] & 1 [1, 1] & 1 [1, 1] \\
 & 0.001 & \textit{0.0000} [0.0000, 0.0003] & 1 [1, 1] & 1 [1, 1] \\
 & 0.01 & \textbf{0.0000} [0.0000, 0.0002] & 1 [1, 1] & 1 [1, 1] \\
 & 0.1 & \textit{0.0000} [0.0000, 0.0002] & 1 [1, 1] & 1 [1, 1] \\
\addlinespace[2pt]
E & 0.0 & 0.0099 [0.0051, 1.7696] & 53 [35, 58] & 4 [4, 4] \\
 & 0.001 & 0.0082 [0.0046, 0.0212] & 28 [19, 40] & 4 [4, 4] \\
 & 0.01 & 0.0125 [0.0051, 0.0548] & 17 [10, 23] & 4 [4, 4] \\
 & 0.1 & 0.0055 [0.0028, 0.0168] & 10 [7, 12] & 4 [4, 4] \\
\addlinespace[2pt]
EF & 0.0 & 0.0031 [0.0011, 0.0112] & 87 [39, 106] & 4 [3, 4] \\
 & 0.001 & \textit{0.0001} [0.0000, 0.0026] & 1 [1, 5] & 1 [1, 2] \\
 & 0.01 & \textit{0.0000} [0.0000, 0.0010] & 1 [1, 2] & 1 [1, 1] \\
 & 0.1 & \textit{0.0000} [0.0000, 0.0010] & 1 [1, 1] & 1 [1, 1] \\
\addlinespace[2pt]
EX & 0.0 & \textit{0.0000} [0.0000, 0.0001] & 1 [1, 1] & 1 [1, 1] \\
 & 0.001 & \textit{0.0000} [0.0000, 0.1898] & 1 [1, 10] & 1 [1, 2] \\
 & 0.01 & \textit{0.0000} [0.0000, 0.0001] & 1 [1, 1] & 1 [1, 1] \\
 & 0.1 & \textit{0.0000} [0.0000, 0.0001] & 1 [1, 1] & 1 [1, 1] \\
\addlinespace[2pt]
EFX & 0.0 & \textit{0.0004} [0.0000, 0.0060] & 11 [1, 106] & 2 [1, 4] \\
 & 0.001 & \textit{0.0001} [0.0000, 0.0142] & 1 [1, 8] & 1 [1, 2] \\
 & 0.01 & \textit{0.0000} [0.0000, 0.0078] & 1 [1, 1] & 1 [1, 1] \\
 & 0.1 & \textit{0.0000} [0.0000, 0.0077] & 1 [1, 1] & 1 [1, 1] \\
\bottomrule
\end{tabular}
\end{minipage}
\qquad
\begin{minipage}[t]{0.35\textwidth}
\centering
\setlength{\tabcolsep}{3pt}
\begin{tabular}{l r r r r}
\toprule
\multicolumn{5}{c}{\textbf{F2: $y = x^4 + x^3 + x^2 + x$}} \\
\midrule
Cond. & $\beta$ & RMSE & Cont.\ Edges & Depth \\
\midrule
Baseline & --- & 0.0012 [0.0007, 0.0062] & 60 [60, 60] & 4 [4, 4] \\
\addlinespace[2pt]
F & --- & \textit{0.0004} [0.0001, 0.0028] & 106 [106, 106] & 4 [4, 4] \\
\addlinespace[2pt]
X & 0.0 & \textit{0.0003} [0.0001, 0.0009] & 10 [1, 10] & 2 [1, 2] \\
 & 0.001 & \textit{0.0001} [0.0001, 0.0005] & 1 [1, 1] & 1 [1, 1] \\
 & 0.01 & \textit{0.0002} [0.0001, 0.0004] & 1 [1, 1] & 1 [1, 1] \\
 & 0.1 & \textit{0.0001} [0.0001, 0.0005] & 1 [1, 1] & 1 [1, 1] \\
\addlinespace[2pt]
FX & 0.0 & \textit{0.0002} [0.0001, 0.0010] & 1 [1, 11] & 1 [1, 2] \\
 & 0.001 & \textit{0.0001} [0.0001, 0.0003] & 1 [1, 1] & 1 [1, 1] \\
 & 0.01 & \textit{0.0001} [0.0001, 0.0002] & 1 [1, 1] & 1 [1, 1] \\
 & 0.1 & \textit{0.0001} [0.0001, 0.0003] & 1 [1, 1] & 1 [1, 1] \\
\addlinespace[2pt]
E & 0.0 & 0.0110 [0.0036, 0.0254] & 53 [40, 60] & 4 [4, 4] \\
 & 0.001 & 0.0103 [0.0027, 0.0441] & 32 [21, 47] & 4 [4, 4] \\
 & 0.01 & 0.0102 [0.0059, 0.0450] & 19 [11, 26] & 4 [4, 4] \\
 & 0.1 & 0.0123 [0.0040, 0.0396] & 11 [9, 13] & 4 [4, 4] \\
\addlinespace[2pt]
EF & 0.0 & 0.0025 [0.0005, 0.0191] & 80 [68, 106] & 4 [4, 4] \\
 & 0.001 & \textit{0.0001} [0.0000, 0.0007] & 1 [1, 7] & 1 [1, 2] \\
 & 0.01 & \textit{0.0001} [0.0000, 0.0091] & 1 [1, 2] & 1 [1, 1] \\
 & 0.1 & \textit{0.0001} [0.0000, 0.0007] & 1 [1, 1] & 1 [1, 1] \\
\addlinespace[2pt]
EX & 0.0 & \textit{0.0001} [0.0000, 0.0002] & 1 [1, 1] & 1 [1, 1] \\
 & 0.001 & \textit{0.0001} [0.0000, 0.0002] & 1 [1, 1] & 1 [1, 1] \\
 & 0.01 & \textit{0.0001} [0.0000, 0.0002] & 1 [1, 1] & 1 [1, 1] \\
 & 0.1 & \textit{0.0001} [0.0000, 0.0002] & 1 [1, 1] & 1 [1, 1] \\
\addlinespace[2pt]
EFX & 0.0 & \textit{0.0010} [0.0005, 0.0135] & 11 [11, 106] & 2 [2, 4] \\
 & 0.001 & \textit{0.0002} [0.0000, 0.0007] & 2 [1, 4] & 1 [1, 2] \\
 & 0.01 & \textit{0.0001} [0.0000, 0.0094] & 1 [1, 1] & 1 [1, 1] \\
 & 0.1 & \textbf{0.0001} [0.0000, 0.0091] & 1 [1, 1] & 1 [1, 1] \\
\bottomrule
\end{tabular}
\end{minipage}
\end{sidewaystable}
\begin{sidewaystable}[p]
\centering
\small
\caption{Absolute results: F3 (left) and F4 (right). Median [min, max] across 10 seeds.}
\label{tab:raw-F03-raw-F04}
\footnotesize
\begin{minipage}[t]{0.35\textwidth}
\centering
\setlength{\tabcolsep}{3pt}
\begin{tabular}{l r r r r}
\toprule
\multicolumn{5}{c}{\textbf{F3: $y = x^5 + x^4 + x^3 + x^2 + x$}} \\
\midrule
Cond. & $\beta$ & RMSE & Cont.\ Edges & Depth \\
\midrule
Baseline & --- & 0.0020 [0.0002, 0.0112] & 60 [60, 60] & 4 [4, 4] \\
\addlinespace[2pt]
F & --- & \textit{0.0003} [0.0001, 0.0018] & 106 [106, 106] & 4 [4, 4] \\
\addlinespace[2pt]
X & 0.0 & \textit{0.0003} [0.0002, 0.0056] & 10 [1, 10] & 2 [1, 2] \\
 & 0.001 & \textit{0.0002} [0.0001, 0.0003] & 1 [1, 1] & 1 [1, 1] \\
 & 0.01 & \textit{0.0002} [0.0001, 0.0003] & 1 [1, 1] & 1 [1, 1] \\
 & 0.1 & \textit{0.0002} [0.0001, 0.0003] & 1 [1, 1] & 1 [1, 1] \\
\addlinespace[2pt]
FX & 0.0 & \textit{0.0002} [0.0001, 0.0008] & 1 [1, 11] & 1 [1, 2] \\
 & 0.001 & \textit{0.0002} [0.0001, 0.0003] & 1 [1, 1] & 1 [1, 1] \\
 & 0.01 & \textit{0.0002} [0.0001, 0.0002] & 1 [1, 1] & 1 [1, 1] \\
 & 0.1 & \textit{0.0002} [0.0001, 0.0003] & 1 [1, 1] & 1 [1, 1] \\
\addlinespace[2pt]
E & 0.0 & 0.0112 [0.0047, 0.0883] & 52 [35, 59] & 4 [4, 4] \\
 & 0.001 & 0.0115 [0.0051, 0.1971] & 31 [19, 40] & 4 [4, 4] \\
 & 0.01 & 0.0082 [0.0045, 0.1099] & 26 [17, 47] & 4 [4, 4] \\
 & 0.1 & 0.0089 [0.0051, 0.0385] & 12 [10, 16] & 4 [4, 4] \\
\addlinespace[2pt]
EF & 0.0 & \textit{0.0017} [0.0003, 0.0179] & 84 [39, 104] & 4 [3, 4] \\
 & 0.001 & \textit{0.0004} [0.0001, 0.7721] & 2 [1, 74] & 2 [1, 4] \\
 & 0.01 & \textit{0.0002} [0.0001, 0.0102] & 2 [1, 3] & 1 [1, 2] \\
 & 0.1 & \textit{0.0002} [0.0001, 0.0047] & 1 [1, 1] & 1 [1, 1] \\
\addlinespace[2pt]
EX & 0.0 & \textit{0.0001} [0.0001, 0.0002] & 1 [1, 1] & 1 [1, 1] \\
 & 0.001 & \textit{0.0002} [0.0001, 0.0678] & 1 [1, 10] & 1 [1, 2] \\
 & 0.01 & \textit{0.0001} [0.0001, 0.0002] & 1 [1, 1] & 1 [1, 1] \\
 & 0.1 & \textit{0.0001} [0.0001, 0.0002] & 1 [1, 1] & 1 [1, 1] \\
\addlinespace[2pt]
EFX & 0.0 & \textit{0.0010} [0.0001, 0.0033] & 10 [1, 11] & 2 [1, 2] \\
 & 0.001 & \textit{0.0003} [0.0001, 0.0013] & 1 [1, 7] & 1 [1, 2] \\
 & 0.01 & \textit{0.0001} [0.0001, 0.0123] & 1 [1, 1] & 1 [1, 1] \\
 & 0.1 & \textbf{0.0001} [0.0001, 0.0117] & 1 [1, 1] & 1 [1, 1] \\
\bottomrule
\end{tabular}
\end{minipage}
\qquad
\begin{minipage}[t]{0.35\textwidth}
\centering
\setlength{\tabcolsep}{3pt}
\begin{tabular}{l r r r r}
\toprule
\multicolumn{5}{c}{\textbf{F4: $y = x^6 + x^5 + x^4 + x^3 + x^2 + x$}} \\
\midrule
Cond. & $\beta$ & RMSE & Cont.\ Edges & Depth \\
\midrule
Baseline & --- & 0.0029 [0.0007, 0.0144] & 60 [60, 60] & 4 [4, 4] \\
\addlinespace[2pt]
F & --- & \textit{0.0011} [0.0002, 0.0028] & 106 [106, 106] & 4 [4, 4] \\
\addlinespace[2pt]
X & 0.0 & \textit{0.0006} [0.0003, 0.0020] & 10 [10, 10] & 2 [2, 2] \\
 & 0.001 & \textit{0.0003} [0.0003, 0.0006] & 1 [1, 1] & 1 [1, 1] \\
 & 0.01 & \textit{0.0003} [0.0003, 0.0006] & 1 [1, 1] & 1 [1, 1] \\
 & 0.1 & \textit{0.0003} [0.0003, 0.0006] & 1 [1, 1] & 1 [1, 1] \\
\addlinespace[2pt]
FX & 0.0 & \textit{0.0004} [0.0003, 0.0056] & 6 [1, 11] & 2 [1, 2] \\
 & 0.001 & \textit{0.0004} [0.0003, 0.0007] & 1 [1, 1] & 1 [1, 1] \\
 & 0.01 & \textit{0.0004} [0.0003, 0.0007] & 1 [1, 1] & 1 [1, 1] \\
 & 0.1 & \textit{0.0004} [0.0003, 0.0007] & 1 [1, 1] & 1 [1, 1] \\
\addlinespace[2pt]
E & 0.0 & 0.0122 [0.0047, 0.0224] & 54 [44, 60] & 4 [4, 4] \\
 & 0.001 & 0.0095 [0.0043, 0.0693] & 27 [21, 50] & 4 [4, 4] \\
 & 0.01 & 0.0088 [0.0036, 0.0131] & 26 [17, 28] & 4 [4, 4] \\
 & 0.1 & 0.0169 [0.0049, 0.0488] & 13 [8, 15] & 4 [4, 4] \\
\addlinespace[2pt]
EF & 0.0 & 0.0064 [0.0030, 0.0211] & 100 [68, 106] & 4 [4, 4] \\
 & 0.001 & \textit{0.0005} [0.0002, 0.0023] & 3 [1, 10] & 2 [1, 3] \\
 & 0.01 & \textit{0.0003} [0.0002, 0.0013] & 1 [1, 1] & 1 [1, 1] \\
 & 0.1 & \textit{0.0003} [0.0002, 0.0011] & 1 [1, 1] & 1 [1, 1] \\
\addlinespace[2pt]
EX & 0.0 & \textit{0.0003} [0.0002, 0.0005] & 1 [1, 1] & 1 [1, 1] \\
 & 0.001 & \textbf{0.0003} [0.0002, 0.0005] & 1 [1, 1] & 1 [1, 1] \\
 & 0.01 & \textit{0.0003} [0.0002, 0.0005] & 1 [1, 1] & 1 [1, 1] \\
 & 0.1 & \textit{0.0003} [0.0002, 0.0005] & 1 [1, 1] & 1 [1, 1] \\
\addlinespace[2pt]
EFX & 0.0 & \textit{0.0021} [0.0006, 0.0265] & 11 [1, 106] & 2 [1, 4] \\
 & 0.001 & \textit{0.0004} [0.0003, 0.0014] & 2 [1, 9] & 2 [1, 2] \\
 & 0.01 & \textit{0.0003} [0.0002, 0.0106] & 1 [1, 3] & 1 [1, 2] \\
 & 0.1 & \textit{0.0003} [0.0002, 0.0127] & 1 [1, 1] & 1 [1, 1] \\
\bottomrule
\end{tabular}
\end{minipage}
\end{sidewaystable}
\begin{sidewaystable}[p]
\centering
\small
\caption{Absolute results: F5 (left) and F6 (right). Median [min, max] across 10 seeds.}
\label{tab:raw-F05-raw-F06}
\footnotesize
\begin{minipage}[t]{0.35\textwidth}
\centering
\setlength{\tabcolsep}{3pt}
\begin{tabular}{l r r r r}
\toprule
\multicolumn{5}{c}{\textbf{F5: $y = \sin(x^2)\cos(x) - 1$}} \\
\midrule
Cond. & $\beta$ & RMSE & Cont.\ Edges & Depth \\
\midrule
Baseline & --- & 0.0003 [0.0001, 0.0011] & 60 [60, 60] & 4 [4, 4] \\
\addlinespace[2pt]
F & --- & \textit{0.0002} [0.0001, 0.0006] & 106 [106, 106] & 4 [4, 4] \\
\addlinespace[2pt]
X & 0.0 & \textit{0.0001} [0.0000, 0.0029] & 6 [1, 10] & 2 [1, 2] \\
 & 0.001 & \textit{0.0001} [0.0000, 0.0001] & 1 [1, 1] & 1 [1, 1] \\
 & 0.01 & \textit{0.0001} [0.0000, 0.0001] & 1 [1, 1] & 1 [1, 1] \\
 & 0.1 & \textit{0.0001} [0.0001, 0.0001] & 1 [1, 1] & 1 [1, 1] \\
\addlinespace[2pt]
FX & 0.0 & \textit{0.0001} [0.0000, 0.0001] & 1 [1, 11] & 1 [1, 2] \\
 & 0.001 & \textit{0.0001} [0.0000, 0.0001] & 1 [1, 1] & 1 [1, 1] \\
 & 0.01 & \textit{0.0001} [0.0000, 0.0003] & 1 [1, 1] & 1 [1, 1] \\
 & 0.1 & \textit{0.0001} [0.0000, 0.0002] & 1 [1, 1] & 1 [1, 1] \\
\addlinespace[2pt]
E & 0.0 & 0.0026 [0.0014, 0.0205] & 58 [51, 60] & 4 [4, 4] \\
 & 0.001 & 0.0017 [0.0007, 0.0098] & 14 [11, 21] & 4 [4, 4] \\
 & 0.01 & 0.0050 [0.0024, 0.0184] & 7 [5, 8] & 4 [4, 4] \\
 & 0.1 & --- & --- & --- \\
\addlinespace[2pt]
EF & 0.0 & 0.0034 [0.0008, 0.0088] & 103 [91, 106] & 4 [4, 4] \\
 & 0.001 & 0.0011 [0.0002, 0.0105] & 2 [2, 4] & 1 [1, 2] \\
 & 0.01 & \textit{0.0001} [0.0000, 0.0081] & 2 [2, 2] & 1 [1, 1] \\
 & 0.1 & \textit{0.0000} [0.0000, 0.0006] & 1 [1, 1] & 1 [1, 1] \\
\addlinespace[2pt]
EX & 0.0 & 0.0009 [0.0000, 0.0115] & 10 [1, 10] & 2 [1, 2] \\
 & 0.001 & 0.0017 [0.0000, 0.0258] & 7 [1, 10] & 2 [1, 2] \\
 & 0.01 & \textit{0.0000} [0.0000, 0.0028] & 1 [1, 5] & 1 [1, 2] \\
 & 0.1 & \textbf{0.0000} [0.0000, 0.0001] & 1 [1, 1] & 1 [1, 1] \\
\addlinespace[2pt]
EFX & 0.0 & 0.0027 [0.0005, 0.0123] & 86 [8, 104] & 4 [2, 4] \\
 & 0.001 & 0.0006 [0.0000, 0.0022] & 2 [2, 2] & 1 [1, 1] \\
 & 0.01 & \textit{0.0001} [0.0000, 0.0096] & 2 [1, 2] & 1 [1, 1] \\
 & 0.1 & \textit{0.0000} [0.0000, 0.0091] & 1 [1, 1] & 1 [1, 1] \\
\bottomrule
\end{tabular}
\end{minipage}
\qquad
\begin{minipage}[t]{0.35\textwidth}
\centering
\setlength{\tabcolsep}{3pt}
\begin{tabular}{l r r r r}
\toprule
\multicolumn{5}{c}{\textbf{F6: $y = \sin(x) + \sin(x + x^2)$}} \\
\midrule
Cond. & $\beta$ & RMSE & Cont.\ Edges & Depth \\
\midrule
Baseline & --- & 0.0007 [0.0003, 0.0027] & 60 [60, 60] & 4 [4, 4] \\
\addlinespace[2pt]
F & --- & \textit{0.0003} [0.0001, 0.0013] & 106 [106, 106] & 4 [4, 4] \\
\addlinespace[2pt]
X & 0.0 & \textit{0.0003} [0.0000, 0.0011] & 10 [1, 10] & 2 [1, 2] \\
 & 0.001 & \textit{0.0001} [0.0001, 0.0002] & 1 [1, 1] & 1 [1, 1] \\
 & 0.01 & \textit{0.0001} [0.0000, 0.0002] & 1 [1, 1] & 1 [1, 1] \\
 & 0.1 & \textit{0.0001} [0.0001, 0.0002] & 1 [1, 1] & 1 [1, 1] \\
\addlinespace[2pt]
FX & 0.0 & \textit{0.0001} [0.0001, 0.0003] & 1 [1, 11] & 1 [1, 2] \\
 & 0.001 & \textit{0.0001} [0.0001, 0.0003] & 1 [1, 1] & 1 [1, 1] \\
 & 0.01 & \textit{0.0001} [0.0001, 0.0004] & 1 [1, 1] & 1 [1, 1] \\
 & 0.1 & \textit{0.0001} [0.0001, 0.0004] & 1 [1, 1] & 1 [1, 1] \\
\addlinespace[2pt]
E & 0.0 & 0.0062 [0.0018, 0.0183] & 51 [37, 60] & 4 [4, 4] \\
 & 0.001 & 0.0067 [0.0021, 0.0136] & 30 [23, 36] & 4 [4, 4] \\
 & 0.01 & 0.0120 [0.0042, 0.0880] & 19 [12, 23] & 4 [4, 4] \\
 & 0.1 & 0.0093 [0.0041, 0.0301] & 8 [6, 10] & 4 [4, 4] \\
\addlinespace[2pt]
EF & 0.0 & 0.0037 [0.0010, 0.0221] & 74 [39, 94] & 4 [3, 4] \\
 & 0.001 & \textit{0.0001} [0.0000, 0.0013] & 2 [1, 7] & 1 [1, 3] \\
 & 0.01 & \textit{0.0001} [0.0000, 0.0014] & 1 [1, 1] & 1 [1, 1] \\
 & 0.1 & \textit{0.0001} [0.0000, 0.0176] & 1 [1, 1] & 1 [1, 1] \\
\addlinespace[2pt]
EX & 0.0 & \textit{0.0001} [0.0000, 0.0001] & 1 [1, 1] & 1 [1, 1] \\
 & 0.001 & \textit{0.0001} [0.0000, 0.0001] & 1 [1, 1] & 1 [1, 1] \\
 & 0.01 & \textit{0.0001} [0.0000, 0.0001] & 1 [1, 1] & 1 [1, 1] \\
 & 0.1 & \textit{0.0001} [0.0000, 0.0001] & 1 [1, 1] & 1 [1, 1] \\
\addlinespace[2pt]
EFX & 0.0 & \textit{0.0004} [0.0000, 0.0032] & 10 [1, 11] & 2 [1, 2] \\
 & 0.001 & \textit{0.0001} [0.0000, 0.0177] & 1 [1, 3] & 1 [1, 2] \\
 & 0.01 & \textbf{0.0001} [0.0000, 0.0041] & 1 [1, 1] & 1 [1, 1] \\
 & 0.1 & \textit{0.0001} [0.0000, 0.0040] & 1 [1, 1] & 1 [1, 1] \\
\bottomrule
\end{tabular}
\end{minipage}
\end{sidewaystable}
\begin{sidewaystable}[p]
\centering
\small
\caption{Absolute results: F7 (left) and F8 (right). Median [min, max] across 10 seeds.}
\label{tab:raw-F07-raw-F08}
\footnotesize
\begin{minipage}[t]{0.35\textwidth}
\centering
\setlength{\tabcolsep}{3pt}
\begin{tabular}{l r r r r}
\toprule
\multicolumn{5}{c}{\textbf{F7: $y = \log(x+1) + \log(x^2+1)$}} \\
\midrule
Cond. & $\beta$ & RMSE & Cont.\ Edges & Depth \\
\midrule
Baseline & --- & 0.0008 [0.0002, 0.0420] & 60 [60, 60] & 4 [4, 4] \\
\addlinespace[2pt]
F & --- & \textit{0.0002} [0.0001, 0.0019] & 106 [106, 106] & 4 [4, 4] \\
\addlinespace[2pt]
X & 0.0 & \textit{0.0002} [0.0000, 0.0046] & 10 [1, 10] & 2 [1, 2] \\
 & 0.001 & \textit{0.0000} [0.0000, 0.0001] & 1 [1, 1] & 1 [1, 1] \\
 & 0.01 & \textit{0.0000} [0.0000, 0.0002] & 1 [1, 1] & 1 [1, 1] \\
 & 0.1 & \textit{0.0000} [0.0000, 0.0003] & 1 [1, 1] & 1 [1, 1] \\
\addlinespace[2pt]
FX & 0.0 & \textbf{0.0000} [0.0000, 0.0007] & 1 [1, 11] & 1 [1, 2] \\
 & 0.001 & \textit{0.0000} [0.0000, 0.0003] & 1 [1, 1] & 1 [1, 1] \\
 & 0.01 & \textit{0.0000} [0.0000, 0.0001] & 1 [1, 1] & 1 [1, 1] \\
 & 0.1 & \textit{0.0000} [0.0000, 0.0001] & 1 [1, 1] & 1 [1, 1] \\
\addlinespace[2pt]
E & 0.0 & 0.0081 [0.0036, 0.0676] & 55 [41, 60] & 4 [4, 4] \\
 & 0.001 & 0.0074 [0.0034, 0.0142] & 28 [16, 32] & 4 [4, 4] \\
 & 0.01 & 0.0040 [0.0022, 0.0091] & 16 [13, 20] & 4 [4, 4] \\
 & 0.1 & 0.0106 [0.0044, 0.0270] & 10 [5, 11] & 4 [4, 4] \\
\addlinespace[2pt]
EF & 0.0 & 0.0046 [0.0030, 0.0104] & 102 [92, 106] & 4 [4, 4] \\
 & 0.001 & 0.0010 [0.0003, 0.0121] & 4 [2, 10] & 2 [1, 2] \\
 & 0.01 & 0.0012 [0.0001, 0.0138] & 2 [2, 4] & 1 [1, 2] \\
 & 0.1 & \textit{0.0001} [0.0000, 0.0019] & 2 [1, 2] & 1 [1, 1] \\
\addlinespace[2pt]
EX & 0.0 & \textit{0.0001} [0.0000, 0.0565] & 1 [1, 10] & 1 [1, 2] \\
 & 0.001 & \textit{0.0000} [0.0000, 0.0782] & 1 [1, 10] & 1 [1, 2] \\
 & 0.01 & \textit{0.0000} [0.0000, 0.0108] & 1 [1, 7] & 1 [1, 2] \\
 & 0.1 & \textit{0.0000} [0.0000, 0.0001] & 1 [1, 1] & 1 [1, 1] \\
\addlinespace[2pt]
EFX & 0.0 & 0.0052 [0.0011, 0.0163] & 103 [11, 106] & 4 [2, 4] \\
 & 0.001 & 0.0010 [0.0004, 0.0125] & 8 [2, 16] & 2 [1, 2] \\
 & 0.01 & \textit{0.0002} [0.0000, 0.0045] & 2 [2, 4] & 1 [1, 2] \\
 & 0.1 & \textit{0.0001} [0.0000, 0.0194] & 1 [1, 2] & 1 [1, 1] \\
\bottomrule
\end{tabular}
\end{minipage}
\qquad
\begin{minipage}[t]{0.35\textwidth}
\centering
\setlength{\tabcolsep}{3pt}
\begin{tabular}{l r r r r}
\toprule
\multicolumn{5}{c}{\textbf{F8: $y = \sqrt{x}$}} \\
\midrule
Cond. & $\beta$ & RMSE & Cont.\ Edges & Depth \\
\midrule
Baseline & --- & 0.0006 [0.0003, 0.0014] & 60 [60, 60] & 4 [4, 4] \\
\addlinespace[2pt]
F & --- & 0.0011 [0.0001, 0.0015] & 106 [106, 106] & 4 [4, 4] \\
\addlinespace[2pt]
X & 0.0 & \textbf{0.0004} [0.0001, 0.0006] & 10 [10, 10] & 2 [2, 2] \\
 & 0.001 & 0.0027 [0.0014, 0.0048] & 1 [1, 1] & 1 [1, 1] \\
 & 0.01 & 0.0027 [0.0014, 0.0048] & 1 [1, 1] & 1 [1, 1] \\
 & 0.1 & 0.0027 [0.0014, 0.0048] & 1 [1, 1] & 1 [1, 1] \\
\addlinespace[2pt]
FX & 0.0 & \textit{0.0005} [0.0001, 0.0014] & 11 [11, 46] & 2 [2, 3] \\
 & 0.001 & 0.0027 [0.0015, 0.0048] & 1 [1, 1] & 1 [1, 1] \\
 & 0.01 & 0.0027 [0.0015, 0.0048] & 1 [1, 1] & 1 [1, 1] \\
 & 0.1 & 0.0027 [0.0015, 0.0048] & 1 [1, 1] & 1 [1, 1] \\
\addlinespace[2pt]
E & 0.0 & 0.0063 [0.0027, 0.0119] & 56 [43, 60] & 4 [4, 4] \\
 & 0.001 & 0.0058 [0.0035, 0.0086] & 26 [15, 30] & 4 [4, 4] \\
 & 0.01 & 0.0044 [0.0022, 0.0079] & 14 [9, 17] & 4 [4, 4] \\
 & 0.1 & 0.0064 [0.0035, 0.1285] & 7 [5, 9] & 4 [4, 4] \\
\addlinespace[2pt]
EF & 0.0 & 0.0063 [0.0019, 0.0127] & 96 [66, 105] & 4 [4, 4] \\
 & 0.001 & 0.0024 [0.0010, 0.0136] & 3 [2, 6] & 2 [1, 2] \\
 & 0.01 & 0.0029 [0.0014, 0.0149] & 2 [2, 2] & 1 [1, 1] \\
 & 0.1 & 0.0027 [0.0014, 0.0102] & 2 [1, 2] & 1 [1, 1] \\
\addlinespace[2pt]
EX & 0.0 & 0.0029 [0.0017, 0.0197] & 10 [1, 34] & 2 [1, 3] \\
 & 0.001 & 0.0028 [0.0015, 0.0097] & 1 [1, 11] & 1 [1, 3] \\
 & 0.01 & 0.0024 [0.0017, 0.0468] & 1 [1, 9] & 1 [1, 2] \\
 & 0.1 & 0.0027 [0.0014, 0.0048] & 1 [1, 1] & 1 [1, 1] \\
\addlinespace[2pt]
EFX & 0.0 & 0.0068 [0.0019, 0.0395] & 92 [8, 105] & 4 [2, 4] \\
 & 0.001 & 0.0021 [0.0006, 0.0050] & 4 [2, 6] & 2 [1, 2] \\
 & 0.01 & 0.0024 [0.0005, 0.0048] & 2 [2, 4] & 1 [1, 2] \\
 & 0.1 & 0.0032 [0.0014, 0.0133] & 1 [1, 2] & 1 [1, 1] \\
\bottomrule
\end{tabular}
\end{minipage}
\end{sidewaystable}
\begin{sidewaystable}[p]
\centering
\small
\caption{Absolute results: F9 (left) and F10 (right). Median [min, max] across 10 seeds.}
\label{tab:raw-F09-raw-F10}
\footnotesize
\begin{minipage}[t]{0.35\textwidth}
\centering
\setlength{\tabcolsep}{3pt}
\begin{tabular}{l r r r r}
\toprule
\multicolumn{5}{c}{\textbf{F9: $z = \sin(x) + \sin(y^2)$}} \\
\midrule
Cond. & $\beta$ & RMSE & Cont.\ Edges & Depth \\
\midrule
Baseline & --- & 0.0010 [0.0004, 0.0058] & 65 [65, 65] & 4 [4, 4] \\
\addlinespace[2pt]
F & --- & \textit{0.0003} [0.0002, 0.0012] & 122 [122, 122] & 4 [4, 4] \\
\addlinespace[2pt]
X & 0.0 & \textit{0.0001} [0.0001, 0.0003] & 2 [2, 2] & 1 [1, 1] \\
 & 0.001 & \textit{0.0001} [0.0001, 0.0002] & 2 [2, 2] & 1 [1, 1] \\
 & 0.01 & \textit{0.0002} [0.0000, 0.0003] & 2 [2, 2] & 1 [1, 1] \\
 & 0.1 & \textit{0.0002} [0.0001, 0.0004] & 2 [2, 2] & 1 [1, 1] \\
\addlinespace[2pt]
FX & 0.0 & \textit{0.0001} [0.0000, 0.0003] & 2 [2, 2] & 1 [1, 1] \\
 & 0.001 & \textit{0.0001} [0.0001, 0.0002] & 2 [2, 2] & 1 [1, 1] \\
 & 0.01 & \textit{0.0001} [0.0001, 0.0003] & 2 [2, 2] & 1 [1, 1] \\
 & 0.1 & \textit{0.0001} [0.0001, 0.0004] & 2 [2, 2] & 1 [1, 1] \\
\addlinespace[2pt]
E & 0.0 & 0.0070 [0.0028, 0.0190] & 64 [52, 65] & 4 [4, 4] \\
 & 0.001 & 0.0037 [0.0022, 0.0136] & 30 [19, 38] & 4 [4, 4] \\
 & 0.01 & 0.0035 [0.0023, 0.0194] & 18 [10, 21] & 4 [4, 4] \\
 & 0.1 & 0.0059 [0.0023, 0.0087] & 7 [6, 12] & 4 [4, 4] \\
\addlinespace[2pt]
EF & 0.0 & 0.0033 [0.0007, 0.0094] & 108 [95, 111] & 4 [4, 4] \\
 & 0.001 & \textit{0.0009} [0.0001, 0.0035] & 4 [2, 6] & 2 [1, 2] \\
 & 0.01 & \textit{0.0005} [0.0000, 0.0026] & 2 [2, 2] & 1 [1, 1] \\
 & 0.1 & \textit{0.0006} [0.0000, 0.0026] & 2 [2, 2] & 1 [1, 1] \\
\addlinespace[2pt]
EX & 0.0 & \textit{0.0001} [0.0000, 0.0107] & 2 [2, 2] & 1 [1, 1] \\
 & 0.001 & \textit{0.0001} [0.0000, 0.0107] & 2 [2, 2] & 1 [1, 1] \\
 & 0.01 & \textit{0.0001} [0.0000, 0.0106] & 2 [2, 2] & 1 [1, 1] \\
 & 0.1 & \textit{0.0001} [0.0000, 0.0106] & 2 [2, 2] & 1 [1, 1] \\
\addlinespace[2pt]
EFX & 0.0 & 0.0038 [0.0014, 0.0110] & 109 [47, 113] & 4 [3, 4] \\
 & 0.001 & \textit{0.0001} [0.0001, 0.0020] & 5 [2, 8] & 2 [1, 2] \\
 & 0.01 & \textbf{0.0001} [0.0000, 0.0020] & 2 [2, 2] & 1 [1, 1] \\
 & 0.1 & \textit{0.0002} [0.0000, 0.0025] & 2 [2, 2] & 1 [1, 1] \\
\bottomrule
\end{tabular}
\end{minipage}
\qquad
\begin{minipage}[t]{0.35\textwidth}
\centering
\setlength{\tabcolsep}{3pt}
\begin{tabular}{l r r r r}
\toprule
\multicolumn{5}{c}{\textbf{F10: $z = 2\sin(x)\cos(y)$}} \\
\midrule
Cond. & $\beta$ & RMSE & Cont.\ Edges & Depth \\
\midrule
Baseline & --- & 0.0295 [0.0089, 0.0637] & 65 [65, 65] & 4 [4, 4] \\
\addlinespace[2pt]
F & --- & \textit{0.0043} [0.0019, 0.0089] & 122 [122, 122] & 4 [4, 4] \\
\addlinespace[2pt]
X & 0.0 & \textit{0.0031} [0.0020, 0.0111] & 15 [15, 40] & 2 [2, 3] \\
 & 0.001 & \textit{0.0024} [0.0018, 0.0042] & 15 [15, 15] & 2 [2, 2] \\
 & 0.01 & \textit{0.0025} [0.0019, 0.0040] & 15 [15, 15] & 2 [2, 2] \\
 & 0.1 & \textit{0.0027} [0.0019, 0.0043] & 15 [15, 15] & 2 [2, 2] \\
\addlinespace[2pt]
FX & 0.0 & \textit{0.0025} [0.0019, 0.0057] & 17 [17, 57] & 2 [2, 3] \\
 & 0.001 & \textit{0.0026} [0.0015, 0.0109] & 17 [17, 17] & 2 [2, 2] \\
 & 0.01 & \textit{0.0026} [0.0018, 0.0046] & 17 [17, 17] & 2 [2, 2] \\
 & 0.1 & \textbf{0.0024} [0.0017, 0.0039] & 17 [17, 17] & 2 [2, 2] \\
\addlinespace[2pt]
E & 0.0 & \textit{0.0174} [0.0099, 0.0415] & 64 [56, 65] & 4 [4, 4] \\
 & 0.001 & \textit{0.0173} [0.0083, 0.0370] & 50 [42, 56] & 4 [4, 4] \\
 & 0.01 & \textit{0.0177} [0.0091, 0.0279] & 33 [27, 41] & 4 [4, 4] \\
 & 0.1 & 0.0360 [0.0114, 0.1767] & 19 [14, 28] & 4 [4, 4] \\
\addlinespace[2pt]
EF & 0.0 & \textit{0.0090} [0.0061, 0.0218] & 102 [80, 119] & 4 [4, 4] \\
 & 0.001 & \textit{0.0091} [0.0060, 0.0144] & 34 [17, 52] & 4 [2, 4] \\
 & 0.01 & \textit{0.0079} [0.0049, 0.0174] & 20 [16, 29] & 3 [2, 4] \\
 & 0.1 & \textit{0.0089} [0.0053, 0.0184] & 13 [7, 18] & 2 [2, 3] \\
\addlinespace[2pt]
EX & 0.0 & \textit{0.0164} [0.0057, 0.0350] & 15 [15, 39] & 2 [2, 3] \\
 & 0.001 & \textit{0.0134} [0.0054, 0.0528] & 15 [15, 15] & 2 [2, 2] \\
 & 0.01 & \textit{0.0149} [0.0056, 0.0624] & 15 [12, 15] & 2 [2, 2] \\
 & 0.1 & \textit{0.0130} [0.0056, 0.0501] & 12 [9, 15] & 2 [2, 2] \\
\addlinespace[2pt]
EFX & 0.0 & \textit{0.0093} [0.0051, 0.0287] & 108 [57, 120] & 4 [3, 4] \\
 & 0.001 & \textit{0.0079} [0.0053, 0.0476] & 43 [29, 47] & 4 [3, 4] \\
 & 0.01 & \textit{0.0102} [0.0048, 0.0162] & 20 [16, 24] & 2 [2, 3] \\
 & 0.1 & \textit{0.0092} [0.0047, 0.0417] & 10 [6, 14] & 2 [2, 2] \\
\bottomrule
\end{tabular}
\end{minipage}
\end{sidewaystable}

\end{document}